\documentclass[10pt,twocolumn,letterpaper]{article}

\usepackage{cvpr}              

\usepackage{lineno}
\definecolor{cvprblue}{rgb}{0.21,0.49,0.74}
\usepackage[pagebackref,breaklinks,colorlinks,allcolors=cvprblue]{hyperref}
\usepackage{multirow}
\usepackage[accsupp]{axessibility} 

\usepackage{algorithm}
\usepackage{algorithmic}

\usepackage{listings}
\usepackage{colortbl}
\usepackage{amsmath}
\usepackage{threeparttable}
\usepackage{wrapfig}
\usepackage{caption}
\usepackage{mathtools}

\usepackage{amsmath,amsfonts,bm}

\def\eqref#1{equation~\ref{#1}}

\def\1{\bm{1}}

\def\vs{{\bm{s}}}

\def\vx{{\bm{x}}}

\DeclareMathAlphabet{\mathsfit}{\encodingdefault}{\sfdefault}{m}{sl}
\SetMathAlphabet{\mathsfit}{bold}{\encodingdefault}{\sfdefault}{bx}{n}

\def\paperID{1086} 
\def\confName{CVPR}
\def\confYear{2025}

\title{Missing Target-Relevant Information Prediction with World Model for Accurate Zero-Shot Composed Image Retrieval}

\author{Yuanmin Tang$^{1,2}$\hspace{0.8em}Jing Yu$^{3}$\thanks{Corresponding author} \hspace{0.8em}Keke Gai$^{4}$\hspace{0.8em}Jiamin Zhuang $^{1,2}$\hspace{0.8em}Gang Xiong$^{1}$\hspace{0.8em}Gaopeng Gou$^{1}$\hspace{0.8em}Qi Wu$^{5}$ \\ 
\textsuperscript{\rm 1}Institute of Information Engineering, Chinese Academy of Sciences \\
\textsuperscript{\rm 2}School of Cyber Security, University of Chinese Academy of Sciences \\
\textsuperscript{\rm 3}School of Information Engineering, Minzu University of China \\
\textsuperscript{\rm 4}Beijing Institute of Technology, 
\textsuperscript{\rm 5}University of Adelaide \\
    \small{\texttt{\{tangyuanmin,zhuangjiamin,gougaopeng,xionggang\}@iie.ac.cn, jing.yu@muc.edu.cn, }}\\
    \small{\texttt{gaikeke@bit.edu.cn,qi.wu01@adelaide.edu.au}}
}

\begin{document}
\maketitle
\begin{abstract}
Zero-Shot Composed Image Retrieval (ZS-CIR) involves diverse tasks with a broad range of visual content manipulation intent across domain, scene, object, and attribute. The key challenge for ZS-CIR tasks is to modify a reference image according to manipulation text to accurately retrieve a target image, especially when the reference image is missing essential target content. In this paper, we propose a novel prediction-based mapping network, named PrediCIR, to adaptively predict the missing target visual content in reference images in the latent space before mapping for accurate ZS-CIR. Specifically, a world view generation module first constructs a source view by omitting certain visual content of a target view, coupled with an action that includes the manipulation intent derived from existing image-caption pairs. Then, a target content prediction module trains a world model as a predictor to adaptively predict the missing visual information guided by user intention in manipulating text at the latent space. The two modules map an image with the predicted relevant information to a pseudo-word token without extra supervision. Our model shows strong generalization ability on six ZS-CIR tasks. It obtains consistent and significant performance boosts ranging from 1.73\% to 4.45\% over the best methods and achieves new state-of-the-art results on ZS-CIR. Our code is available at \url{https://github.com/Pter61/predicir}.
\end{abstract}    
\section{Introduction}
\label{sec:intro}

\begin{figure}[t]
    \centering
    \includegraphics[width=1.0\linewidth]{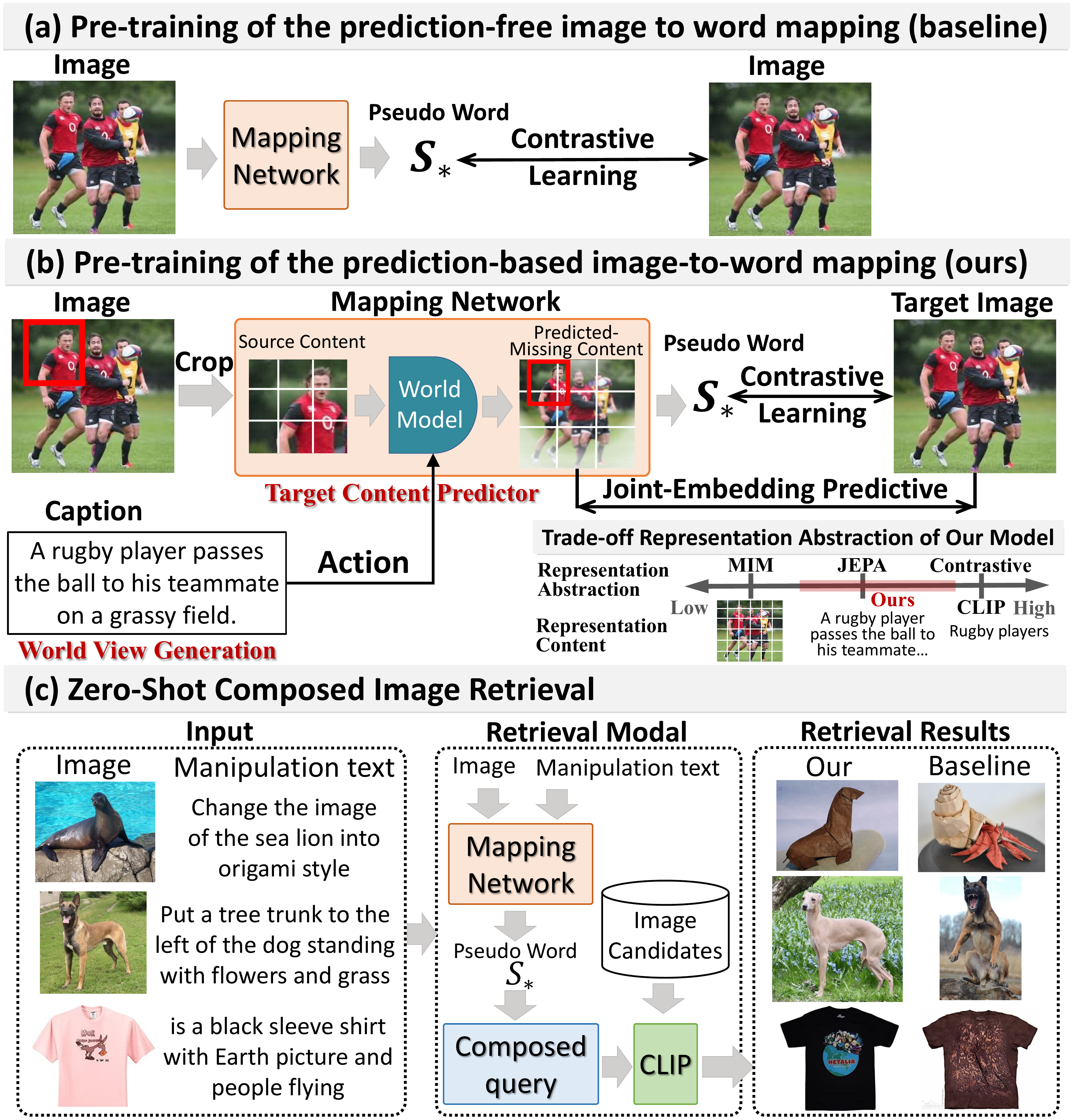}
    \caption{Illustration of our motivation. (a) Prediction-free visual mapping. (b) Our prediction-based visual mapping. (c) ZS-CIR process results from different strategies.} 
    \label{fig:motivation}
    \vspace{-10pt}
\end{figure}

Given a reference image and a human manipulation text, Composed Image Retrieval (CIR) \cite{Vo_2019_CVPR} aims to retrieve a target image visually similar to the reference image while incorporating content modifications specified by the manipulation text.  Distinct from traditional content-based image retrieval \cite{datta2008image}, in the CIR task, the content of the manipulation text often does not appear in the reference images, as illustrated in Figure \ref{fig:motivation}(c). CIR enhances flexibility and improves the accuracy of intent expression by allowing users to integrate visual and textual information into their search queries. This task has gained emerging attention in internet searches and e-commerce applications \cite{Chen_2020_CVPR,Saito_2023_CVPR}. As shown in Figure \ref{fig:motivation}(c), CIR tasks include image domain transformation for creative searches, object composition and manipulation for natural image searches, and attribute modifications for fashion image searches.

 There exist two core challenges of CIR: (1) accurately modify the reference image guided by the manipulation text for retrieving the target image, particularly when the necessary visual content is missing in the reference images, and (2) adaptively compose visual and textual content guided by manipulation text for image retrieval. Various supervised \cite{Liu_2021_ICCV,Baldrati_2022_CVPR} and semi-supervised methods \cite{jang2024visual,gu2023compodiff} have been proposed to address CIR problem, which requires an extensive amount of annotated triplets, \textit{i.e.,} a reference image, a manipulated text, and a target image, for training task-specific retrieval models.  These supervised methods are labor-intensive or require large models (\textit{e.g.,} diffusion \cite{rombach2022high}) for data annotation, which limits their generalizability. To overcome this issue, recent studies have introduced Zero-Shot Composed Image Retrieval (ZS-CIR) \cite{Saito_2023_CVPR,baldrati2023zero} by utilizing a pre-trained CLIP model \cite{radford2021learning} to treat ZS-CIR as a traditional text-based image retrieval challenge.  As depicted in Figure \ref{fig:motivation}(c), these methods map the reference image into a pseudo-token of CLIP's language space, combining it with manipulation text to form a query. This query retrieves target images from a shared semantic space in a zero-shot mode by calculating semantic similarity. Despite these advancements, the mapping networks are inadequate for ZS-CIR tasks for the following reasons:

(1) The CLIP embedding, learned through contrastive methods, is coarse-grained~\cite{garrido2024learningleveragingworldmodels}. This leads to losing crucial visual details in the pre-trained pseudo-token, which is essential for CIR tasks. For instance, in the bottom right of Figure \ref{fig:motivation}(b), the CLIP embedding captures the main object (\textit{e.g.,} rugby players) while missing fine-grained details of relation (\textit{e.g.,} pass ball) and sense (\textit{e.g.,} grassy field).

(2) Existing ZS-CIR methods train a network to map reference images into language space, ignoring its missing target content, as shown in Figure  \ref{fig:motivation}(a). This limits the model's ability to generate target image information for retrieval adaptively. In fact, key elements of the target images are often missing from the reference images. Considering the queries in Figure \ref{fig:motivation}(c), existing methods struggle to handle manipulations where relevant elements are missing in the reference images, such as image domains (\textit{e.g.}, origami style),  objects/scene (\textit{e.g.}, dog among the flowers with tree), and attributes (\textit{e.g.}, black with Earth logo).

In this paper, we propose a novel approach to \textit{\textbf{Predi}ct target image feature before retrieval for zero-shot \textbf{C}omposed \textbf{I}mage \textbf{R}etrieval} (\textbf{PrediCIR}). Unlike existing ZS-CIR approaches, PrediCIR trains a world model, an effective target content predictor \cite{yang2023learning,garrido2024learningleveragingworldmodels}, to adaptively predict key visual elements (\textit{e.g.,} objects, senses, attributes, domain) missing from reference images, guided by manipulation text, as illustrated in Figure \ref{fig:motivation}(b): the \textit{World View Generation} module first generates both source and target views from existing image-caption pairs without extra supervision for training. Specifically, we corrupt an image’s content via random cropping, regarding these as the source view and the original image as the target view, with the corresponding description as the action. Subsequently, the \textit{Target Content Predictor} module trains a world model to adaptively predict the key element of the target view guided by action, which is missing in the source view. The two modules map an image with the predicted target content to a pseudo-word, providing a high-quality feature with fine-grained visual details.

The main contributions are summarized as follows: (1) We introduce a novel prediction-based image-to-word mapping network, leveraging the world model to simulate user behavior. This facilitates the prediction of potential target image features relevant to manipulation intent for retrieval, offering new insights into the vision-to-language alignment mechanism. (2) The proposed mapping network of PrediCIR adaptively predicts key elements (\textit{e.g.,} objects, senses, attributes, and various details) of target visual content missing in reference images, proving advantageous for challenges such as object combination, foreground/background editing, attribute adjustment, and domain conversion. (3) Our PrediCIR is consistently effective for diverse ZS-CIR tasks. It significantly improves CIR from 1.73\% to 4.45\% across six CIR tasks with comparable inference times. It establishes new state-of-the-art results and further impacts a broader range of vision and language applications. 
\section{Related works}
\label{sec:related_works}

\noindent\textbf{Composed Image Retrieval.} Composed Image Retrieval (CIR) combines image and text features for retrieval \cite{Vo_2019_CVPR}, typically using late fusion to integrate visual and language features separately while requiring extensively annotated triplets CIR datasets. \cite{Baldrati_2022_CVPR, Liu_2021_ICCV, jang2024visual, zhang2024magiclens}. Zero-shot CIR models \cite{Saito_2023_CVPR, baldrati2023zero, tang2023contexti2w, gu2024lincir, karthik2024visionbylanguage,du2024image2sentence,jang2024spherical,yang2024semantic,tang2024reason,tang2024denoise}, trained on image-text pairs, avoid the need for extensive CIR datasets by mapping reference images to text space for query formation. However, they often miss visual content specified by manipulation text, resulting in less accurate queries.  To address this, we introduce a prediction-based word mapping, allowing the text encoder to access potential target image features. Unlike CompoDiff \cite{gu2023compodiff}, which requires multi-step diffusion model training with synthesizing triplets, our model predicts target content in latent space on a single step, enhancing performance without additional supervision.  We create pseudo triplets by cropping visual elements to preserve full contextual embeddings under a frozen CLIP, avoiding the mask-based CIR methods \cite{chen2023pretrain,hou2024pseudo,zhang2024zero} that require CLIP fine-tuning.
Additionally, unlike diffusion \cite{gu2023compodiff}, LLMs \cite{karthik2024visionbylanguage}, or external databases \cite{Suo_2024_CVPR,FTI4CIR} methods, which introduce non-negligible computational overhead, our model remains lightweight with comparable inference times.

\begin{figure*}
    \centering
    \includegraphics[width=1.0\linewidth]{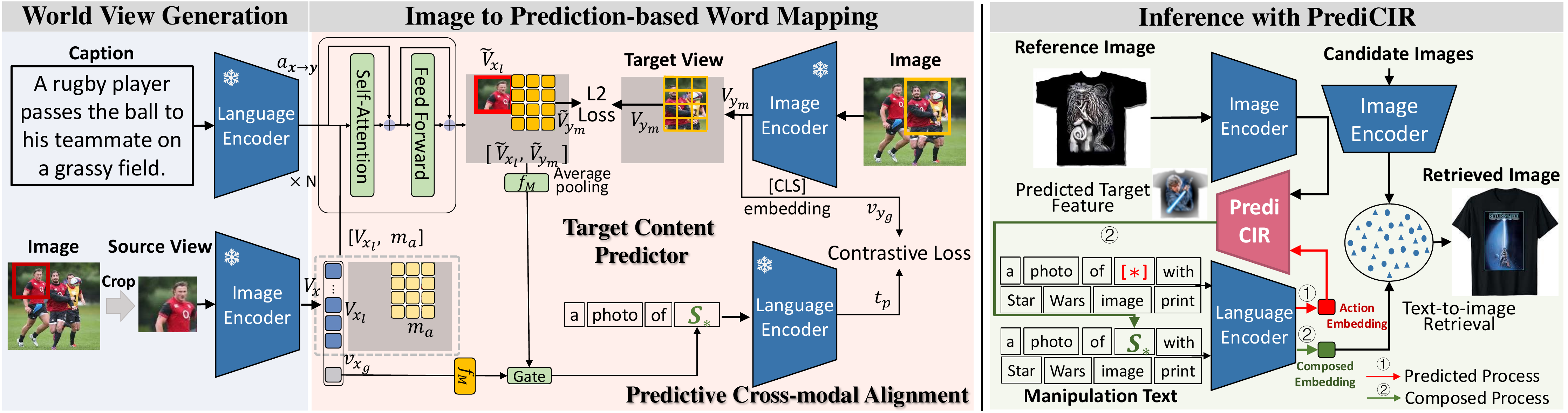}   
    \caption{An overview of our PrediCIR model. Pre-training (left): Image to prediction-based word mapping aims to predict target-relevant missing visual content in latent space and map it with reference image content to a pseudo-word token $S_*$. Inference (right): Map the inference image to $S_*$ and form the composed query in a unified language space for ZS-CIR.}
    \label{fig:model-architecture}
    \vspace{-12pt}
\end{figure*}

\noindent\textbf{World Model in Vision Representation Learning.}  World modeling has proven its effectiveness across several domains \cite{hafner2019dream,yang2023learning}, showing clear benefits in prediction-based representation learning.  Masked Image Modeling (MIM) approaches \cite{He_2022_CVPR, Xie_2022_CVPR} learn representations by predicting masked image areas in pixel space, aligning their decoders with world models. Similarly, leveraging the Joint Embedding Predictive Architecture (JEPA) \cite{lecun2022path}, I-JEPA \cite{assran2023self} predicts masked parts of the image in the latent space. Recent work \cite{garrido2024learningleveragingworldmodels} introduces an image world model that predicts cropped image region in latent space, focusing on fine-grained visual details at the patch level instead of whole-image diffusion \cite{rombach2022high}.   Building on these advances, our work first introduces a novel world model to predict target image features guided by text for vision-language retrieval tasks.

\noindent\textbf{Vision and Language Pre-training Models.}  Vision and Language Pre-training (VLP) models, such as CLIP \cite{radford2021learning}, employ extensive image-text pair training to achieve implicit alignment between visual and textual data. Recent advancements in VLP \cite{Zhou_2022_CVPR, song2022clip} have utilized static models to merge encoded image and text features, facilitating a range of zero-shot tasks \cite{pmlr-v162-li22n, NEURIPS2022_960a172b, li2023blip2, song2022clip, shi2023dual}. In our work, we adapt VLP models for CIR tasks in a zero-shot manner. Moreover, we uniquely leverage the pre-trained knowledge of VLP models to encode world views and actions to train a vision-language world model for prediction-based ZS-CIR.

\section{Methodology}
\label{sec:method}

Given a reference image \( I \) and a manipulation text \( T \), Zero-Shot Composed Image Retrieval (ZS-CIR) aims to retrieve images from an image database that are visually similar to \( I \) while incorporating the visual modifications specified in \( T \). A detailed illustration of our model is provided in Figure \ref{fig:model-architecture}.  We first introduce a new approach for generating world view from image-caption pairs. This process allows us to provide source and target views with corresponding actions for training a predictor based on the Joint Embedding Predictive Architecture (JEPA)~\cite{lecun2022path}. In this framework, the predictor is the instantiation of the world model \cite{assran2023self, garrido2024learningleveragingworldmodels}. Then, we learn a prediction-based mapping network of PrediCIR to predict visual elements of the target image missing in the reference image guided by manipulation text \( T \) and convert them into a pseudo-word token \( S_{*} \) in the token embedding space.  In this work, \( S_{*} \) depicts the potential visual content of the target image, combining the existing content of the reference image with the predicted visual elements specified by \( T \).  To effectively compose  $I$ and $T$ across different modalities for zero-shot image retrieval,  we construct a composed query in the form of a sentence $P$ ``a photo of $S_{*}$, {$T$}'' and embed it using the frozen text encoder of CLIP. Given the composed query embedding, we embed each candidate image $I_{i}$ by the frozen image encoder of CLIP and approach ZS-CIR as a traditional text-to-image retrieval task by measuring the similarity between $P$ and $I_{i}$.

\subsection{World View Generation}

Since images and captions in existing ZS-CIR training datasets share the contextual elements, training a world model to predict the absent visual elements in reference images poses a challenge.  To address this, we construct images with missing visual elements by modifying existing image-caption datasets, specifically by randomly cropping existing images. This approach aligns better with the frozen CLIP model than the masked-region CIR method \cite{chen2023pretrain}, which requires fine-tuning to interpret masked inputs.  Given an image \( I \) with width \( W \) and height \( H \) in an image-caption pair, we obtain a cropped image \( I_c \) as follows:
\begin{equation}
\resizebox{0.909\hsize}{!}{$\begin{aligned}
I_c &= \operatorname{Crop}(I, (x, y, W_c, H_c)) \\
    &= \operatorname{Crop}(I, (x, y, x + \sqrt{s_c r_c WH}, y + \sqrt{s_c WH / r_c}))
\end{aligned}$}
\label{f:crop}
\end{equation}
where \(\operatorname{Crop}(\cdot)\) denotes the cropping operation, and \( (x, y) \) denotes the top-left corner of the cropping region. These coordinates are dynamically calculated to focus on areas of interest within the image. \( W_c \) and \( H_c \) are the width and height of the cropped image, respectively. Crop size \( s_c \) and aspect ratios \( r_c \) ensure the cropping operation dynamically adjusts to various image sizes. Subsequently, we utilize the cropped image as the source view \( \boldsymbol{x} \), the original image as the target view \( \boldsymbol{y} \), and the caption as the action \( a_{\boldsymbol{x}{\rightarrow \boldsymbol{y}}} \) for training a world model for predicting the visual content of a target view that missing in a source view.

As CLIP shows its strong capabilities, we employ the CLIP model to encode the source/target views and the actions for prediction-based mapping. We utilize the visual encoder of the frozen CLIP model to represent the cropped image \( I_c \) as the source view \( \boldsymbol{x} \) by a set of visual feature vectors \( \boldsymbol{V}_x = \{\boldsymbol{v}_{x_{i}}\}_{i=1}^m \) where \( m = 257 \) and \( d = 1024 \). Here, the \( \boldsymbol{v}_{x_{1}} \) denotes the global source feature \( \boldsymbol{v}_{x_{g}} \), and the subsequent vectors \( \{\boldsymbol{v}_{x_{i}}\}_{i=2}^m \) represent the local patch features \( \boldsymbol{V}_{x_{l}} \). Similarly, the original image \( I \) is encoded as the target view \( \boldsymbol{y} \) using a set of visual feature vectors \( \boldsymbol{V}_y = \{\boldsymbol{v}_{y_{i}}\}_{i=1}^m \), where \( \boldsymbol{v}_{y_{1}} \) acting as the global target feature \( \boldsymbol{v}_{y_{g}} \).

In this work,  we construct a dataset of triplets \( <  \)source view, action, target view\( > \) comprising \( <  \)cropped image, caption, original image\( > \) for training the PrediCIR network. Subsequently, we utilize triplets \( <  \)reference image, manipulation text, target image\( > \) triplets for ZS-CIR. We treat both captions and manipulation texts as actions \( a_{\boldsymbol{x}{\rightarrow \boldsymbol{y}}} \), which express the user's intent to modify the source view (\textit{e.g.,} a rugby player) into the target view (\textit{e.g.,} a rugby player pass the ball to his teammate on a grassy field) as illustrated in Figure \ref{fig:model-architecture}.  Our PrediCIR has two goals: First, it predicts the visual content of the target view missing in the source, guided by the action. Second, it adaptively combines the predicted content with the source's content for mapping. To this end, we feed the caption to the frozen CLIP language encoder, obtaining the \texttt{[CLS]} token embedding \( \boldsymbol{t} = \{t_i\}_{i=1}^d \in \mathbb{R}^{d \times 1} \) as the action for predicting.

\subsection{Image to Prediction-based Word Mapping}

Given the constructed triplets \( <  \)source view, action, target view\( > \), where the source view \( \boldsymbol{x} = \boldsymbol{V}_{x_{l}} \) comprises the local patch-level features of the cropped image, the target view \( \boldsymbol{y} = \boldsymbol{V}_y \) includes the patch-level features of the original image, and the action \( a_{\boldsymbol{x}{\rightarrow \boldsymbol{y}}} = \boldsymbol{t} \) encapsulates the user manipulation intent. We introduce two modules to progressively predict the target visual content that missing in the source view, and map to a pseudo-token for accurate ZS-CIR: the \textit{Target Content Predictor} (TCP for short) first learns a world model functioning as a user simulator to predict the target visual content guided by the manipulation intent through the JEPA framework. Subsequently, the \textit{Predictive Cross-Modal Alignment} (PMA for short)  adaptively combines the predicted and source visual contents, mapping them into the word token space using cross-modal contrastive learning.

\noindent\textbf{Target Content Predictor.} Given the visual patch features from the triplets \( <  \)source view, action, target view\( > \). This module aims to predict the target visual content missing in the source view, guided by the action. Specifically,  we fed with the target visual content in the form of mask tokens as well as \( a_{\boldsymbol{x}{\rightarrow \boldsymbol{y}}} \). We denote these mask tokens as \(m_a\), parameterized by a shared learnable vector with an added positional embedding, representing a randomly sampled block \( B \) from the target patch features \( \boldsymbol{V}_y \). These mask tokens \(m_a\) corresponding of the position of \( B \) by the target patch features \(\boldsymbol{V}_{y_m} = \{\boldsymbol{V}_{y_{j}}\}_{j \in B}\).  Specifically, we sample the block using the same crop size and aspect ratios described in Eq.\ref{f:crop}. Subsequently, as illustrated in Figure \ref{fig:model-architecture} (left), we apply self-attention and combine the action \( a_{\boldsymbol{x}{\rightarrow \boldsymbol{y}}}\), the embedded source patches \( \boldsymbol{V}_{x_{l}} \), and mask tokens \(m_a\) as input \( \boldsymbol{X} = [a_{\boldsymbol{x}{\rightarrow \boldsymbol{y}}},\boldsymbol{V}_{x_{l}}, m_a]\). First, we compute the query, key and value through linear projections, \textit{i.e.,} $\boldsymbol{Q} = \boldsymbol{X}\boldsymbol{W}^Q$, $\boldsymbol{K} = \boldsymbol{X}\boldsymbol{W}^K$, $\boldsymbol{V} = \boldsymbol{X}\boldsymbol{W}^V$.  $\boldsymbol{X}$ denotes concatenating the three matrices, which enhances the interaction between mask tokens and source local patches guided by the manipulation intent to achieve a high-quality representation with fine-gained visual details crucial for CIR tasks. Then, the mask token and source local patches from the current self-attention block $\boldsymbol{X}^{i}$ are calculated as:
\begin{gather}
\boldsymbol{X}_{att}^{i} = \operatorname{Att}(\boldsymbol{Q}, \boldsymbol{K}, \boldsymbol{V})=\textit{softmax}\left(\frac{\boldsymbol{Q K}^{\top}}{\sqrt{d}}\right)\boldsymbol{V} \\
\boldsymbol{X}^{i} = \operatorname{FFW}(\boldsymbol{X}_{att}^{i} + \boldsymbol{X}^{i-1}) + \boldsymbol{X}_{att}^{i}
\label{f:attn}
\end{gather}
\noindent where $\boldsymbol{X}^{i-1}$ are mask tokens with source local patch features from the previous block and $\operatorname{FFW}(\cdot)$ denotes 2-layer feed-forward networks.  Finally, we calculate the squared \(\mathcal{L}_2\) loss to minimize the distance between the patch prediction  \(\Tilde{\boldsymbol{V}}_{y_m} = \boldsymbol{X}_{out}[\operatorname{select}(m_a)]\) as follow: 
\begin{equation}
    \mathcal{L}_{pred} = \mathcal{L}_2(\boldsymbol{x}, \boldsymbol{y}) = \sum_{i\in B}\| \Tilde{\boldsymbol{V}}^{i}_{y_{m}} 
    - \boldsymbol{V}^{i}_{y_{m}} \|_2^2
\end{equation}
\noindent where $\boldsymbol{X}_{out}$ denotes the output embeddings from $N$ transformer blocks. \(\operatorname{select}(\cdot) \) is used to select the corresponding indexes of features within $\boldsymbol{X}_{out}$, \( \Tilde{\boldsymbol{V}}^{i}_{y_{m}}\) and \(\boldsymbol{V}^{i}_{y_{m}}\) are the \(i^{\text{th}}\) target patch prediction and the corresponding target patch feature, respectively, and \( B\) is the target image block.

\begin{table*}[t]
\centering
\scalebox{0.95}
{\footnotesize
\setlength{\tabcolsep}{3.5mm}
\begin{tabular}{ccccccccccccc}
\toprule
                            &                         &                                 & \multicolumn{2}{c}{Dress}                                                               & \multicolumn{2}{c}{Shrit}                                                      & \multicolumn{2}{c}{TopTee}                                                              & \multicolumn{2}{c}{Average}                               \\ \cmidrule(lr){4-5}\cmidrule(lr){6-7}\cmidrule(lr){8-9}\cmidrule(lr){10-11}
Backbones                 & Methods               & Conferences                                   & R10                                  & R50                                              & R10                         & R50                                              & R10                                  & R50                                              & R10                         & R50                         \\ \cmidrule(lr){1-11}
                            & Pic2Word$^{\dagger}$ & CVPR 2023                                                & 20.0                                 & \multicolumn{1}{c|}{40.2}                        & 26.2                        & \multicolumn{1}{c|}{43.6}                        & 27.9                                 & \multicolumn{1}{c|}{47.4}                        & 24.7                        & 43.7                        \\
                            & SEARLE-XL$^{\dagger}$ & ICCV 2023                                                & 20.3                                 & \multicolumn{1}{c|}{43.2}                        & 27.4                                 & \multicolumn{1}{c|}{45.7}                        & 29.3                                 & \multicolumn{1}{c|}{50.2}                        & 25.7                        & 46.3                        \\
                            & LinCIR$^{\dagger}$ & CVPR 2024                                                & 20.9                                 & \multicolumn{1}{c|}{42.4}                        & 29.1                        & \multicolumn{1}{c|}{46.8}                        & 28.8                                 & \multicolumn{1}{c|}{50.2}                        & 26.3                        & 46.5                        \\ 
                            & Context-I2W$^{\dagger}$ & AAAI 2024                                              & 23.1                        & \multicolumn{1}{c|}{45.3}               & 29.7                        & \multicolumn{1}{c|}{48.6}               & 30.6                        & \multicolumn{1}{c|}{52.9}               & 27.8              & 48.9 \\
                            \multirow{-5}{*}{ViT-L/14} & \textbf{PrediCIR} & --                                              & \textbf{25.4}                        & \multicolumn{1}{c|}{\textbf{49.5}}               & \textbf{31.8}                        & \multicolumn{1}{c|}{\textbf{52.0}}               & \textbf{33.1}                        & \multicolumn{1}{c|}{\textbf{55.4}}               & \textbf{30.1}              & \textbf{52.3} \\ \cmidrule(lr){1-11}
                            & CompoDiff$^{\dagger}$ & TMLR 2024                                                & 37.8                                  & \multicolumn{1}{c|}{49.1}                        & 41.3                                 & \multicolumn{1}{c|}{55.2}                        &  44.3                                 & \multicolumn{1}{c|}{56.4}                        &     39.0                    &    51.7                     \\
                            & LinCIR$^{\dagger}$ & CVPR 2024                                                & 38.1                                 & \multicolumn{1}{c|}{60.9}                        & 46.8                                & \multicolumn{1}{c|}{65.1}                        &  50.5                                 & \multicolumn{1}{c|}{71.1}                        &     45.1                    &    65.7                     \\
                            \multirow{-3}{*}{ViT-G/14} & \textbf{PrediCIR} & --                                              & \textbf{39.7}                        & \multicolumn{1}{c|}{\textbf{62.4}}               & \textbf{48.2}                        & \multicolumn{1}{c|}{\textbf{67.4}}               & \textbf{53.7}                        & \multicolumn{1}{c|}{\textbf{73.6}}               & \textbf{47.2}              & \textbf{67.8} \\

                            \bottomrule
\end{tabular}}
\caption{Results on Fashion-IQ for attribute manipulation. $^{\dagger}$indicates results from the original paper.}
\label{tab:fashion}
\vspace{-10pt}
\end{table*}

\noindent\textbf{Predictive Cross-Modal Alignment.} Given the target patch prediction  \( \Tilde{\boldsymbol{V}}_{y_m} \), enhanced source patch features \( \Tilde{\boldsymbol{V}}_{x_l}=\boldsymbol{X}_{out}[\operatorname{select}(x_l)]\) that is high-quality representation, and global source feature \( \boldsymbol{v}_{x_{g}} \), the AI agent aims to form a target embedding optimized for retrieval. When mapping the visual content to a pseudo-word token, both the predict and source content are complementary to form the complete target information.  To align the feature of JEPA and CLIP and adaptively weight predicted information on the retrieval process, we introduce a learnable scalar \( gate \) that decides the contribution of the predicted information  \( [\Tilde{\boldsymbol{V}}_{x_l}, \Tilde{\boldsymbol{V}}_{y_m}] \) and integrates the global source information \( \boldsymbol{v}_{x_{g}} \) to form the final target embedding $\boldsymbol{S}_{*}$ as follows:  
\begin{equation}
S_{*} = f_{M_p}(gate\cdot\operatorname{Avg}([\Tilde{\boldsymbol{V}}_{x_l}, \Tilde{\boldsymbol{V}}_{y_m}]))+f_{M_s}(\boldsymbol{v}_{x_{g}}) 
\label{f:gate}
\end{equation}
\noindent where $\operatorname{Avg}(\cdot)$ denotes average pooling, $f_{M_p}(\cdot)$ and $f_{M_s}(\cdot)$ respectively denote predict and source mapping of 3-layer feed-forward networks. To map the pseudo token $S_{*}$ to the word token space, we append \(S_{*}\) to the end of the token embeddings of the prompted sentence, ``\texttt{a photo of}'', and feed it to the language encoder of CLIP to obtain the sentence embedding $\boldsymbol{t}_p$. We aim to match an image to its paired prediction-based prompt sentence while separating unpaired ones. We minimize the symmetric contrastive loss between the global target visual feature \( \boldsymbol{v}_{y_g} \) and the prompt sentence embedding \( \boldsymbol{t}_p \) as follows:
\begin{equation}
\begin{aligned}
\mathcal{L}_{align}=\mathcal{L}_{t 2 i}(\boldsymbol{t}_p,\boldsymbol{v}_{y_g})+\mathcal{L}_{i 2 t}(\boldsymbol{t}_p, \boldsymbol{v}_{y_g})
\end{aligned}
\label{f:loss}
\end{equation}
The two contrastive loss terms with a temperature hyper-parameter $\tau$ that controls the strength of penalties on hard negative samples are defined as:
\begin{equation}
\begin{aligned}
\mathcal{L}_{ t 2 i}(\boldsymbol{t}_p, \boldsymbol{v}_{y_g}) = -\frac{1}{|\mathcal{B}|} \sum_{i \in \mathcal{B}} \log \frac{e^{\tau (\boldsymbol{t}_p^{i})^T \boldsymbol{v}_{y_g}^{i}}}{\sum_{j \in \mathcal{B}} e^{\tau (\boldsymbol{t}_p^{i})^T \boldsymbol{v}_{y_g}^{j}}}
\end{aligned}
\label{f:CTL_1}
\end{equation}
\begin{equation}
\begin{aligned}
\mathcal{L}_{ i 2 t}(\boldsymbol{t}_p, \boldsymbol{v}_{y_g}) = -\frac{1}{|\mathcal{B}|} \sum_{i \in \mathcal{B}} \log \frac{e^{\tau (\boldsymbol{v}_{y_g}^{i})^T \boldsymbol{t}_p^{i}}}{\sum_{j \in \mathcal{B}} e^{\tau (\boldsymbol{v}_{y_g}^{i})^T \boldsymbol{t}_p^{j}}}
\end{aligned}
\label{f:CTL_2}
\end{equation}

\noindent The ﬁnal loss to optimize PrediCIR:
\begin{equation}
\begin{aligned}
\mathcal{L}=\mathcal{L}_{pred} + \mathcal{L}_{align}
\end{aligned}
\label{f:train_loss}
\end{equation}

\subsection{Inference with PrediCIR}
In the inference stage, we compose the reference image with the paired manipulation text and compare the composed query with candidate images for retrieval. As shown in Figure \ref{fig:model-architecture} (right), we first form a prompt sentence that includes special token \texttt{[*]} and manipulation text, which is fed to the language encoder of CLIP to obtain an action embedding, followed by predicting through the PrediCIR network. Then, we obtain the mapped pseudo-token embedding $S_*$ containing the predicted target-relevant information and replace the \texttt{[*]} token with $S_*$ to form a composed query. The result is embedded by the language encoder and compared to the visual features of candidate images. 

Since our focus is on studying the prediction-based word mapping for ZS-CIR, we utilize the same prompt in the most recent works \cite{Saito_2023_CVPR,tang2023contexti2w} for a fair comparison. We show prompt examples for different ZS-CIR tasks. In all examples, \texttt{[*]} indicates the pseudo token from the mapping network: \textbf{(a) Domain conversion} aims to modify the domain of the reference image. The prompt is defined as \texttt{a [domain tag] of [*]}; \textbf{(b) Object composition} retrieves an image that contains an object in the reference image and other object tags. The prompt is in the format of \texttt{a photo of [*], [obj$_1$ tag] and [obj$_2$ tag], $\dots$, and [obj$_n$ tag]}; \textbf{(c) Sentence manipulation} modifies the reference image based on a sentence. We simply append the sentence with the special token as  \texttt{a photo of [*], [sentence]}.

\section{Experiments}
\label{sec:exp}

\begin{figure}[t]
    \centering
    \centering
    \includegraphics[width=1.0\linewidth]{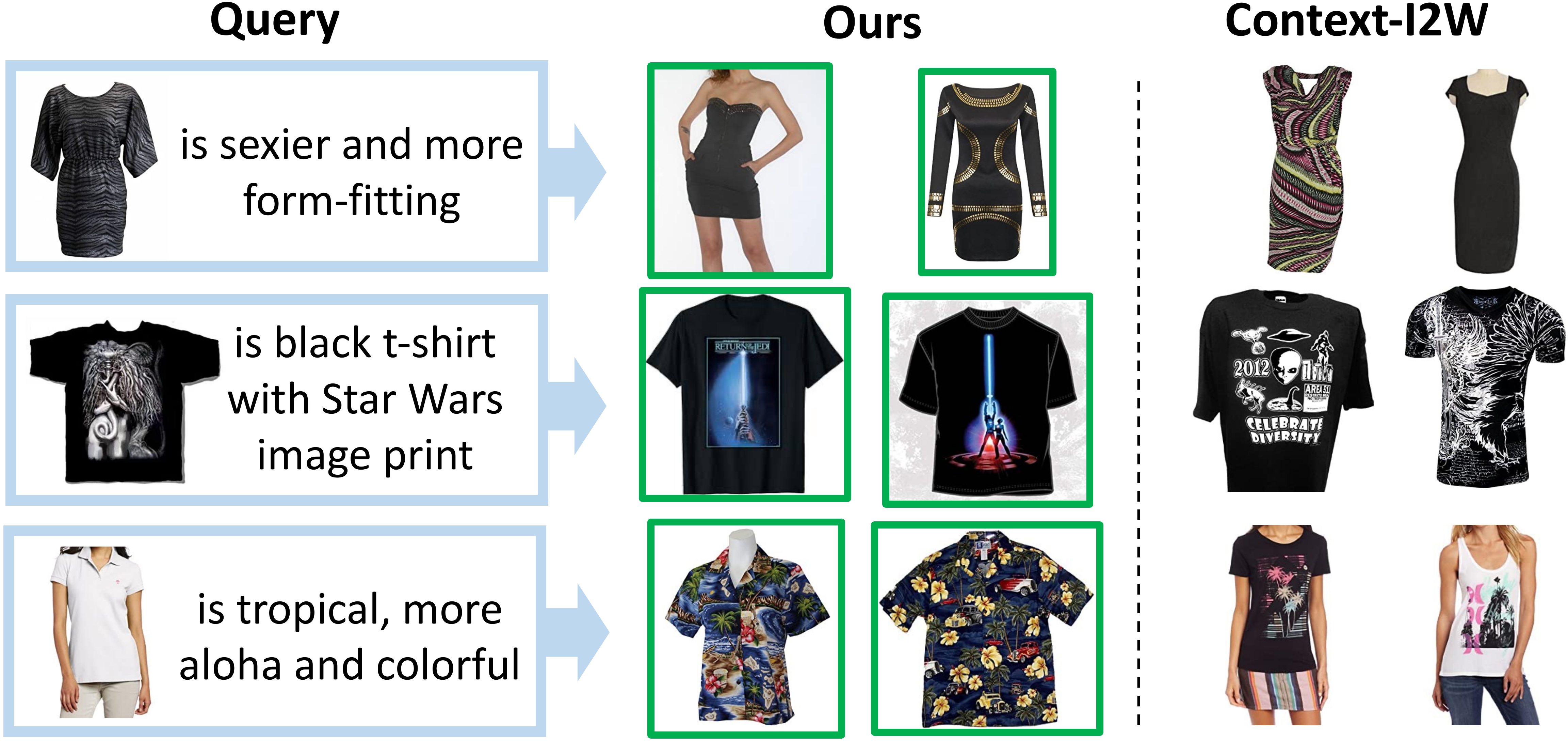}
    \caption{Results of attribute manipulation on FashionIQ.}
    \label{fig:fashion}
    \vspace{-10pt}
\end{figure}

\noindent \textbf{Datasets.} We evaluate our model on four ZS-CIR datasets, \textit{i.e.,} COCO \cite{10.1007/978-3-319-10602-1_48} and GeneCIS \cite{vaze2023genecis} for object/attribute composition, ImageNet \cite{deng2009imagenet,Hendrycks_2021_ICCV} for domain conversion, CIRR \cite{Liu_2021_ICCV} and CIRCO \cite{baldrati2023zero} for object/scene manipulation, and Fashion-IQ \cite{Wu_2021_CVPR} for attribute manipulation. All the dataset settings and evaluation metrics (Recall@K and mAP@R) follow the recent works \cite{Saito_2023_CVPR,baldrati2023zero,gu2024lincir} for a fair comparison. 

(1) Domain conversion.  This dataset comprises 16,983 images of 200 classes from four domains, \textit{i.e.,} cartoon, origami, toy, and sculpture. We use the prompt (a) in inference. \noindent (2) Object/attribute composition. The COCO dataset contains images with corresponding lists of object labels and instance masks of query images. Similarly, the GeneCIS dataset introduces four task variations, such as changing a specific attribute or object. We use the prompt (b) in inference. \noindent (3) Object/scene manipulation. A reference image is an instruction for manipulating an object or the background. We apply the prompt (c) in inference. \noindent (4) Attribute manipulation.  This dataset includes various descriptions for manipulating image attributes. We utilize the prompt (c) in inference. Details are in the Appendix \ref{secc::more_inference}.

\noindent \textbf{Implementation Details.} We adopt ViT-L/14 CLIP \cite{radford2021learning} from OpenAI and ViT-G/14 CLIP \cite{openclip} from OpenCLIP. The crop sizes of random cropped images and blocked target images are the same in the $(0.2, 0.25)$, and the aspect ratios are $(0.75, 1.5)$, respectively. For training PrediCIR, we utilize the Conceptual Caption dataset \cite{DBLP:conf/acl/SoricutDSG18}, which comprises 3M images.  The number of self-attention blocks is $12$ with $384$ dimensional embeddings. To improve training stability, we initialize the learnable scalar of tanh-gating to 0 \cite{bachlechner2021rezero}. We employ AdamW \cite{loshchilov2018decoupled} with a learning rate of $1\times10^{-5}$, weight decay of $0.1$, and a linear warmup of $10000$ steps. The batch size is $1024$.  All models are trained on \(4\) NVIDIA A100 (80G) GPUs. To ensure reliable results, we report the performance averaged over three trials.

\begin{table}[t]
\centering
\scalebox{0.95}
{\footnotesize
\setlength{\tabcolsep}{3.5mm}
\begin{tabular}{ccccccccccc}
\toprule
Backbones                                      & Methods                                                   & R1                                               & R5                                               & \multicolumn{1}{l}{R10}                                              \\ \midrule
\multicolumn{1}{c|}{}                            & \multicolumn{1}{c|}{Pic2Word$^{\dagger}$}                             & 23.9                                            & 51.7                                           & 65.3                                                                                  \\
\multicolumn{1}{c|}{}                            & \multicolumn{1}{c|}{SEARLE-XL$^{\dagger}$}                   & 24.2                                            & 52.4                                            & 66.3                                                                             \\
\multicolumn{1}{c|}{}                            & \multicolumn{1}{c|}{LinCIR$^{\dagger}$}                   & 25.0                                            & 53.3                                            & 66.7                                                                            \\
\multicolumn{1}{c|}{}                           & \multicolumn{1}{c|}{Context-I2W$^{\dagger}$}                         & 25.6                                   & 55.1                                   & 68.5                                                                      \\
\multicolumn{1}{c|}{\multirow{-5}{*}{ViT-L/14}}                             & \multicolumn{1}{c|}{\textbf{PrediCIR}}                         & \textbf{27.2}                                   & \textbf{57.0}                                   & \textbf{70.2}                                                                      \\ \midrule
\multicolumn{1}{c|}{}                            & \multicolumn{1}{c|}{CompoDiff$^{\dagger}$}                   & 26.7                                             & 55.1                                             & 74.5                                                                       \\
\multicolumn{1}{c|}{}                            & \multicolumn{1}{c|}{LinCIR$^{\dagger}$}                   & 35.3                                            & 64.7                                            & 76.1                                                                       \\
\multicolumn{1}{c|}{\multirow{-3}{*}{ViT-G/14}}                             & \multicolumn{1}{c|}{\textbf{PrediCIR}}                         & \textbf{37.0}                                   & \textbf{66.1}                                   & \textbf{77.9}                                                                    \\  
\bottomrule
\end{tabular}}
\caption{Results on CIRR for object manipulation.}
\label{tab:CIRR}
\vspace{-10pt}
\end{table}

\begin{table}[t]
\centering
\scalebox{0.95}
{\footnotesize
\setlength{\tabcolsep}{1.0mm}
\begin{tabular}{ccccccccccc}
\toprule
Backbones                                      & Methods                                                   & mAP@5                      & mAP@10                         & mAP@25                                               & \multicolumn{1}{l}{mAP@50}                                              \\ \midrule
\multicolumn{1}{c|}{}                            & \multicolumn{1}{c|}{Pic2Word}                             & 8.7 & 9.5 & 10.6 & 11.3                                                                                  \\
\multicolumn{1}{c|}{}                            & \multicolumn{1}{c|}{SEARLE-XL$^{\dagger}$}                   & 11.7 & 12.7 & 14.3 & 15.1                                                                             \\
\multicolumn{1}{c|}{}                            & \multicolumn{1}{c|}{LinCIR$^{\dagger}$}                  & 12.6  & 13.6  & 15.0  & 15.9                                                                            \\
\multicolumn{1}{c|}{}                           & \multicolumn{1}{c|}{Context-I2W}                         & 13.0  & 14.6  & 16.1  & 17.2                                                                      \\
\multicolumn{1}{c|}{\multirow{-5}{*}{ViT-L/14}}                             & \multicolumn{1}{c|}{\textbf{PrediCIR}}                         & \textbf{15.7}                                   & \textbf{17.1}                 &   \textbf{18.6}                & \textbf{19.3}                                                                      \\ \midrule
\multicolumn{1}{c|}{}                            & \multicolumn{1}{c|}{CompoDiff$^{\dagger}$}                   & 15.3 & 17.7 & 19.5 & 21.0                                                                       \\
\multicolumn{1}{c|}{}                            & \multicolumn{1}{c|}{LinCIR$^{\dagger}$}                   & 19.7 & 21.0 & 23.1 & 24.2                                                                       \\
\multicolumn{1}{c|}{\multirow{-3}{*}{ViT-G/14}}                             & \multicolumn{1}{c|}{\textbf{PrediCIR}}                         & \textbf{23.7}                                   & \textbf{24.6}            &   \textbf{25.4}                     & \textbf{26.0}                                                                    \\  
\bottomrule
\end{tabular}}
\caption{Results on CIRCO for object manipulation.}
\label{tab:CIRCO}
\vspace{-10pt}
\end{table}

\begin{figure}[t]
    \centering
    \centering
    \includegraphics[width=1.0\linewidth]{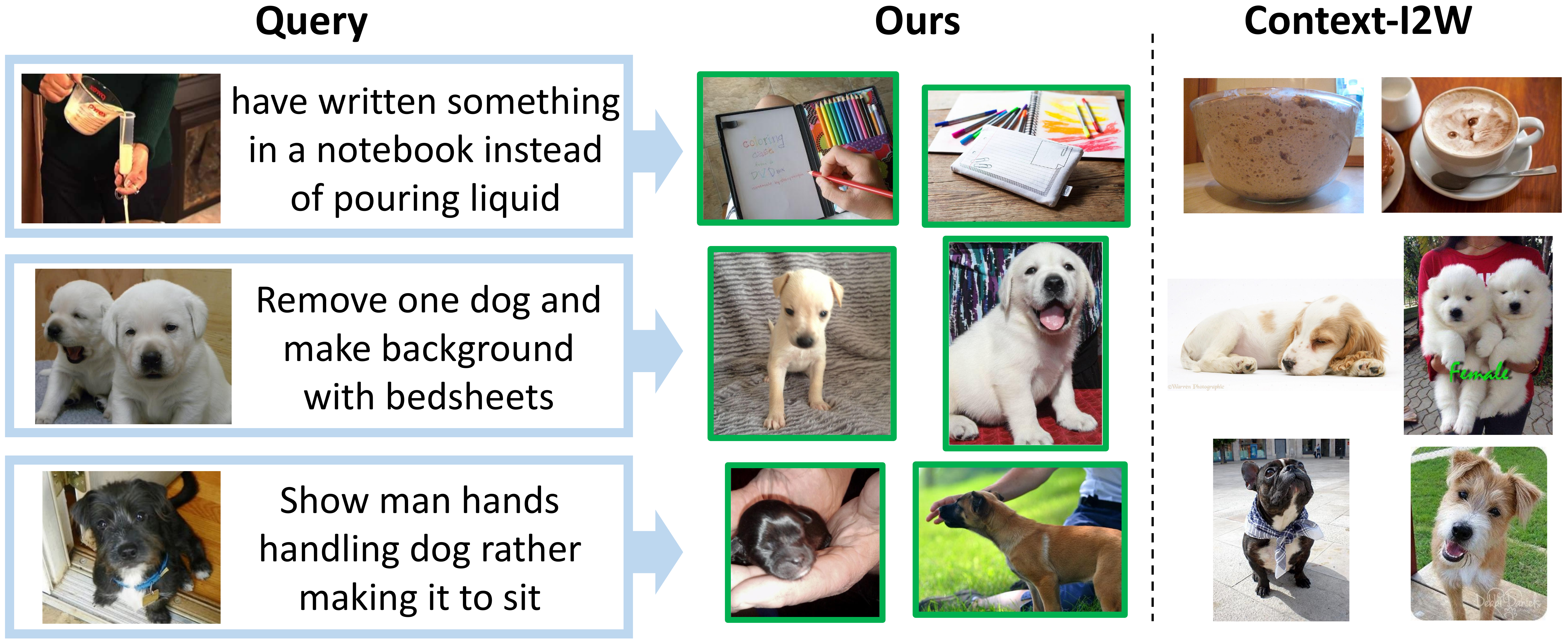}
    \caption{Results of the object manipulation on CIRR.}
    \label{fig:cirr}
    \vspace{-10pt}
\end{figure}

\begin{table}[t]
\centering
\scalebox{0.95}
{\footnotesize
\setlength{\tabcolsep}{3.5mm}
\begin{tabular}{ccccccccccc}
\toprule
Backbones                                      & Methods                                                   & R1                                               & R5                                               & \multicolumn{1}{l}{R10}                                              \\ \midrule
\multicolumn{1}{c|}{}                            & \multicolumn{1}{c|}{Pic2Word$^{\dagger}$}                            & 11.5          & 24.8          & 33.4                                                                                  \\
\multicolumn{1}{c|}{}                            & \multicolumn{1}{c|}{SEARLE-XL}                   & 13.3                                            & 28.3                                            & 37.6                                                                             \\
\multicolumn{1}{c|}{}                            & \multicolumn{1}{c|}{LinCIR}                   & 11.7                                            & 24.9                                            & 34.2                                                                            \\
\multicolumn{1}{c|}{}                           & \multicolumn{1}{c|}{Context-I2W$^{\dagger}$}                         & 13.5          & 28.5          & 38.1                                                                      \\
\multicolumn{1}{c|}{\multirow{-5}{*}{ViT-L/14}}                             & \multicolumn{1}{c|}{\textbf{PrediCIR}}                         & \textbf{15.1}                                   & \textbf{33.0}                                   & \textbf{42.8}                                                                      \\ \midrule
\multicolumn{1}{c|}{}                            & \multicolumn{1}{c|}{LinCIR}                   & 14.8           & 30.6          & 40.5                                                                                    \\
\multicolumn{1}{c|}{\multirow{-2}{*}{ViT-G/14}}                             & \multicolumn{1}{c|}{\textbf{PrediCIR}}                         & \textbf{17.2}                                   & \textbf{34.8}                                   & \textbf{45.9}                                                                    \\  
\bottomrule
\end{tabular}}
\caption{Results on COCO for object composition.}
\label{tab:coco}
\vspace{-10pt}

\end{table}

\begin{table*}[ht]
\centering
\scalebox{0.85}
{\footnotesize
\setlength{\tabcolsep}{3.5mm}
\begin{tabular}{ccccccccccccc}
\toprule
                           &       &       & \multicolumn{2}{c}{Cartoon}                       & \multicolumn{2}{c}{Origami}                        & \multicolumn{2}{c}{Toy}                            & \multicolumn{2}{c}{Sculpture}                      & \multicolumn{2}{c}{Average}   \\  \cmidrule(lr){4-5}\cmidrule(lr){6-7}\cmidrule(lr){8-9}\cmidrule(lr){10-11}\cmidrule(lr){12-13}
Backbones                & Methods   & Conferences    & R10          & R50                                & R10           & R50                                & R10           & R50                                & R10           & R50                                & R10           & R50           \\ \midrule
\multirow{5}{*}{ViT-L/14}  & Pic2Word$^{\dagger}$ & CVPR 2023    & 8.0          & \multicolumn{1}{c|}{21.9}          & 13.5          & \multicolumn{1}{c|}{25.6}          & 8.7           & \multicolumn{1}{c|}{21.6}          & 10.0          & \multicolumn{1}{c|}{23.8}          & 10.1          & 23.2          \\ 
                          & SEARLE-XL & ICCV 2023 & 9.6 & \multicolumn{1}{c|}{24.9} & 16.1 & \multicolumn{1}{c|}{27.3} & 7.6 & \multicolumn{1}{c|}{25.4} & 11.3 & \multicolumn{1}{c|}{26.4} & 11.2 & 26.0 \\
                            & LinCIR & CVPR 2024 & 9.4 & \multicolumn{1}{c|}{24.2} & 15.7 & \multicolumn{1}{c|}{26.9} & 10.8 & \multicolumn{1}{c|}{27.0} & 11.7 & \multicolumn{1}{c|}{27.9} & 11.9 & 26.5 \\
                           & Context-I2W$^{\dagger}$ & AAAI 2024 & 10.2 & \multicolumn{1}{c|}{26.1} & 17.5 & \multicolumn{1}{c|}{28.7} & 11.6 & \multicolumn{1}{c|}{27.4} & 12.1 & \multicolumn{1}{c|}{28.2} & 12.9 & 27.6 \\
                           & \textbf{PrediCIR} & -- & \textbf{14.2} & \multicolumn{1}{c|}{\textbf{31.9}} & \textbf{20.4} & \multicolumn{1}{c|}{\textbf{34.3}} & \textbf{14.7} & \multicolumn{1}{c|}{\textbf{30.8}} & \textbf{16.3} & \multicolumn{1}{c|}{\textbf{34.9}} & \textbf{16.4} & \textbf{33.0} \\ \midrule
                           \multirow{2}{*}{ViT-G/14}  & LinCIR & CVPR 2024 & 13.7          & \multicolumn{1}{c|}{30.2}          & 19.5          & \multicolumn{1}{c|}{32.9}          & 13.8           & \multicolumn{1}{c|}{30.2}          & 15.2          & \multicolumn{1}{c|}{34.0}          & 15.5        &  31.8     \\
& \textbf{PrediCIR} & -- & \textbf{15.6} & \multicolumn{1}{c|}{\textbf{34.6}} & \textbf{23.7} & \multicolumn{1}{c|}{\textbf{37.2}} & \textbf{17.2} & \multicolumn{1}{c|}{\textbf{37.5}} & \textbf{19.3} & \multicolumn{1}{c|}{\textbf{37.8}} & \textbf{19.0} & \textbf{36.8} \\  
\bottomrule
\end{tabular}}
\caption{Results on ImageNet for domain conversion. $^{\dagger}$indicates results from the original paper.}
\label{tab:imgnet}
\vspace{-15pt}
\end{table*}

\subsection{Quantitative and Qualitative Results}
We compare PrediCIR with several commonly benchmarked ZS-CIR methods, including: 1) \textbf{Pic2Word} \cite{Saito_2023_CVPR}: maps a reference image into a pseudo-word token within the CLIP token embedding space. 2) \textbf{SEARLE} \cite{baldrati2023zero}: Integrates the pseudo-word token with the GPT caption \cite{brown2020language}. 3) \textbf{Context-I2W} \cite{tang2023contexti2w}: Selectively extracts text description-relevant visual information before mapping.  4) \textbf{LinCIR} \cite{gu2024lincir}: Masks subjects in captions for efficiency training. For a fair comparison, we present the results of methods relying on the ViT-L/14 and ViT-G/14 CLIP models without LLMs \cite{karthik2024visionbylanguage} or external databases \cite{Suo_2024_CVPR,FTI4CIR}.  We also compare with the semi-supervised 5) \textbf{CompoDiff} \cite{gu2023compodiff}: Training a diffusion model using 18M synthetic data for multi-step entire target image prediction.  We report results for CompoDiff on ViT-G/14 CLIP, given its comparable inference times. Since most baselines reported their results on ViT-L/14, we primarily compare results on this backbone and explore the generalization ability of our model on ViT-G/14.

\begin{figure}[t]
    \centering
    \centering
    \includegraphics[width=1.0\linewidth]{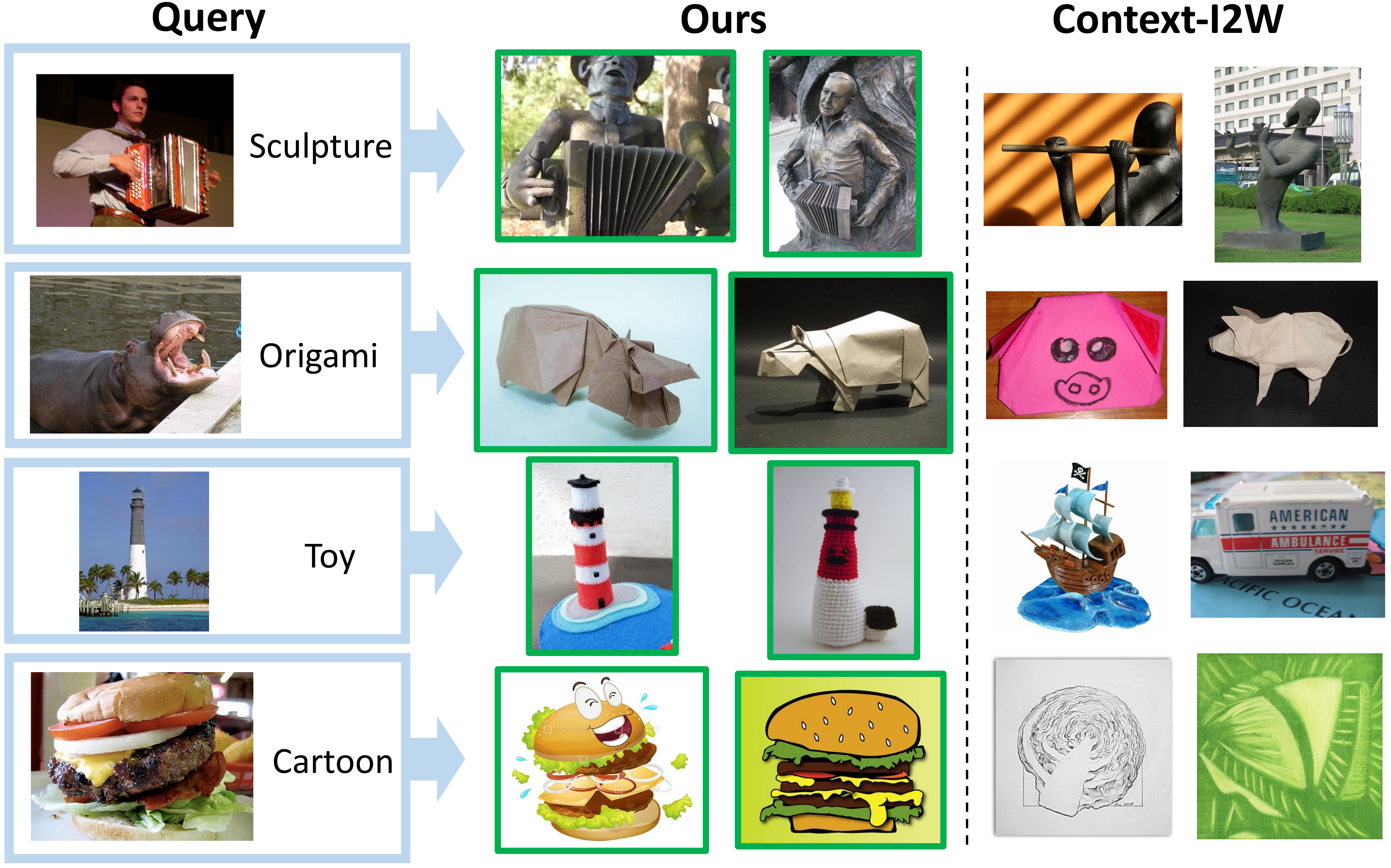}
    \caption{Retrieved results on the domain conversion task.}
    \label{fig:imgnet}
    \vspace{-10pt}
    
\end{figure}

PrediCIR surpasses existing ZS-CIR models on the ViT-L/14 and ViT-G/14 backbones. Tables \ref{tab:fashion} to \ref{tab:genecls} present the quantitative results, while Figures \ref{fig:fashion} to \ref{fig:imgnet} illustrate the corresponding qualitative results of our model and the most recent works, Context-I2W.  The attribute manipulation task requires accurately localizing specific attributes within the fashion image. As indicated in Table \ref{tab:fashion}, PrediCIR achieves an average improvement of 2.85\% on ViT-L/14, over the State-of-the-Art (SoTA) model, Context-I2W. Context-I2W struggles to retrieve a target image with accurately manipulated fashion attributes, which are missing in the reference images. PrediCIR tackles this challenge by effectively predicting fashion-relevant visual details guided by manipulation text for CIR retrieval. As exemplified in Figure \ref{fig:fashion}, PrediCIR accurately predicts the missing fashion-relevant attribute of sexier and form-fitting (row 1), Star Wars print (row 2), and tropical with more aloha and colorful (row 3).

We further evaluate PrediCIRs' capability in foreground/background differentiation and fine-grained image editing through the object/scene manipulation task (Table \ref{tab:CIRR} and Table \ref{tab:CIRCO}). PrediCIR consistently surpasses existing ZS-CIR models, achieving an average performance improvement of 1.73\% over the best model on CIRR and 2.45\% on CIRCO. This improvement is attributed to PrediCIRs' approach of predicting target elements that are missing in reference images guided by manipulation intention before searching and mapping into a pseudo token with fine-grained visual details, enhancing the ability of the CLIP language encoder to compose target image information accurately. In Figure \ref{fig:cirr}, PrediCIR accurately predicts the absent fine-grained visual content of a written notebook (row 1), a bedsheet background (row 2), and a handing hand (row 3).

In the object/attribute composition experiments (Table \ref{tab:coco} and \ref{tab:genecls}), PrediCIR significantly outperforms the current SoTA model by an average of 3.60\% on COCO and 3.43\% on GeneCLS. These results underscore the remarkable effectiveness of our TCP module in predicting missing objects relevant to manipulation text.

\begin{table}[t]
\centering
\scalebox{0.95}
{\footnotesize
\setlength{\tabcolsep}{3.5mm}
\begin{tabular}{ccccccccccc}
\toprule
Backbones                                      & Methods                                                   & R1                                               & R2                                               & \multicolumn{1}{l}{R3}                                              \\ \midrule
\multicolumn{1}{c|}{}                            & \multicolumn{1}{c|}{Pic2Word$^{\dagger}$}                            & 11.2           & 21.5         & 30.4                                                                                  \\
\multicolumn{1}{c|}{}                            & \multicolumn{1}{c|}{SEARLE-XL}                   & 12.3                                            & 22.1                                             &  31.3                                                                            \\
\multicolumn{1}{c|}{}                            & \multicolumn{1}{c|}{LinCIR}                   & 12.2                                           & 22.8                                              & 32.4                                                                           \\
\multicolumn{1}{c|}{}                           & \multicolumn{1}{c|}{Context-I2W}                         & 12.5          & 23.2          & 33.1                                                                      \\
\multicolumn{1}{c|}{\multirow{-5}{*}{ViT-L/14}}                             & \multicolumn{1}{c|}{\textbf{PrediCIR}}                         & \textbf{16.6}                                   & \textbf{26.7}                                   & \textbf{35.8}                                                                      \\ \midrule
\multicolumn{1}{c|}{}                            & \multicolumn{1}{c|}{LinCIR$^{\dagger}$}                   & 13.7           & 24.6         & 34.1                                                                                     \\
\multicolumn{1}{c|}{}                            & \multicolumn{1}{c|}{CompoDiff$^{\dagger}$}                   & 15.5           & 26.6         & 35.4                                                                                     \\
\multicolumn{1}{c|}{\multirow{-3}{*}{ViT-G/14}}                             & \multicolumn{1}{c|}{\textbf{PrediCIR}}                         & \textbf{17.7}                                   & \textbf{28.9}                                   & \textbf{38.6}                                                                    \\  
\bottomrule
\end{tabular}}
\caption{Results on GeneCLS. The average R@1, R@2, R@3 for ``Focus Attribute'', ``Change Attribute'', ``Focus Object'', and ``Change Object'' are shown. The full table in Appendix \ref{sec::more_genecis}.}
\label{tab:genecls}
\vspace{-10pt}

\end{table}

Moreover, in the domain conversion experiments (Table \ref{tab:imgnet}), PrediCIR consistently outperforms existing approaches and notably surpasses the SoTA Context-I2W by an average of 4.45\%. As illustrated in Figure \ref{fig:imgnet}, PrediCIR accurately converts image domains guided by manipulation text while maintaining fidelity to the visual content of the reference image  (\textit{e.g.,} man playing accordion, a hippo with mouth open, the lighthouse on the island, and juicy burger).  In contrast, Context-I2W struggles to map images to other domains as specified by manipulation texts while missing fine-grained details in the contrastive representation.
\begin{table}[t]
\centering
\scalebox{0.95}
{\footnotesize
\setlength{\tabcolsep}{1.8mm}
\begin{tabular}{llcclcc}
\toprule
\multicolumn{2}{l}{}   & \multicolumn{3}{c}{CIRR}                         & \multicolumn{2}{c}{Fashion-IQ} \\ \cmidrule(lr){3-5}\cmidrule(lr){6-7}
   & Methods           & R1   & R5   & R10                                & R10            & R50           \\ \midrule
1. & full model        & 27.2 & 57.0 & \multicolumn{1}{l|}{70.2}          & 30.1           & 52.3          \\
2. & w/o cropped images & 23.5  & 53.6 & \multicolumn{1}{l|}{66.0}          & 25.1           & 45.5          \\
3. & w/o action & 22.4 & 52.7 & \multicolumn{1}{l|}{64.9}          & 24.5           & 43.2          \\
4. & w/o Target Predictor             & 20.2 & 44.5 & \multicolumn{1}{l|}{56.3}           & 22.5           & 41.9          \\
5. & w/o \(\mathcal{L}_2\) loss       & 22.0 & 51.5 & \multicolumn{1}{c|}{65.7}          & 24.2           & 42.8          \\
6. & w/o gate          & 25.9 & 55.4 & \multicolumn{1}{l|}{67.8}          & 27.5           & 49.8          \\
7. &  self-attention    & 18.2 & 42.4 & \multicolumn{1}{l|}{55.8}           & 21.3           & 40.5          \\  
8. &  mask images   & 22.3 & 52.2 & \multicolumn{1}{l|}{64.3}          & 24.2           & 42.8          \\
9. & predict entire images  & 25.1 & 54.3 & \multicolumn{1}{l|}{66.8}          & 26.5           & 49.0          \\ 
\bottomrule
\end{tabular}}
\caption{Ablation study on CIRR and FashionIQ.}
\label{tab:ablation}
\vspace{-10pt}
\end{table}

\subsection{Ablation Study}

Following \cite{Saito_2023_CVPR, baldrati2023zero, tang2023contexti2w}, we evaluate the contribution of the core components in PrediCIR with ViT-L/14 backbone on CIRR and Fashion-IQ in Table \ref{tab:ablation}. \textbf{(1) In models `2-3', we evaluate the importance of the world view generation approach.}  Using the entire target image as the source view without cropping images (model `2'), the performance significantly declined by an average of 4.62\%, indicating that a corrupt original image as the source image is essential for learning the ability to predict the target visual content that is missing in the reference image. When removing the action embedding  \( a_{\boldsymbol{x}{\rightarrow \boldsymbol{y}}} \) (model '3') results in a significant drop of 5.82\% on average. \textbf{(2) In models `4-6', we assess the importance of key modules in the prediction-based image-to-word mapping process.} Removing  PTC (model `4') or JEPA framework (model `5') causes obvious performance decrease of 10.28\% and 6.12\% on average, respectively. By directly summing the predicted and original image features instead of using the gating strategy in PMA (model '6'), the performance drops by 2.08\%. It indicates the necessity to capture complementary information from the two sources adaptively. \textbf{(3) Models `7-10' evaluate the effect of alternative solutions for key modules.} In model `7', we replace the PrediCIR with a typical self-attention network with the same input. The results drop significantly by 11.76\% on average, confirming the effectiveness of the PrediCIR mapping strategy. In model `8', we employ random masking for constructing source views. The results drop by 6.20\% on average, likely due to the frozen CLIP encoder struggling with masked images, whereas our cropped images retain coherent regional context. In model `9', we predict the entire target image, resulting in an average drop of 3.02\%, indicating that partial prediction reduces computation and mitigates overfitting. Due to space constraints, please refer to Appendix \ref{secc:more_ablation} for more ablation studies.

\begin{figure}[t]
    \centering
    \centering
    \includegraphics[width=0.95\linewidth]{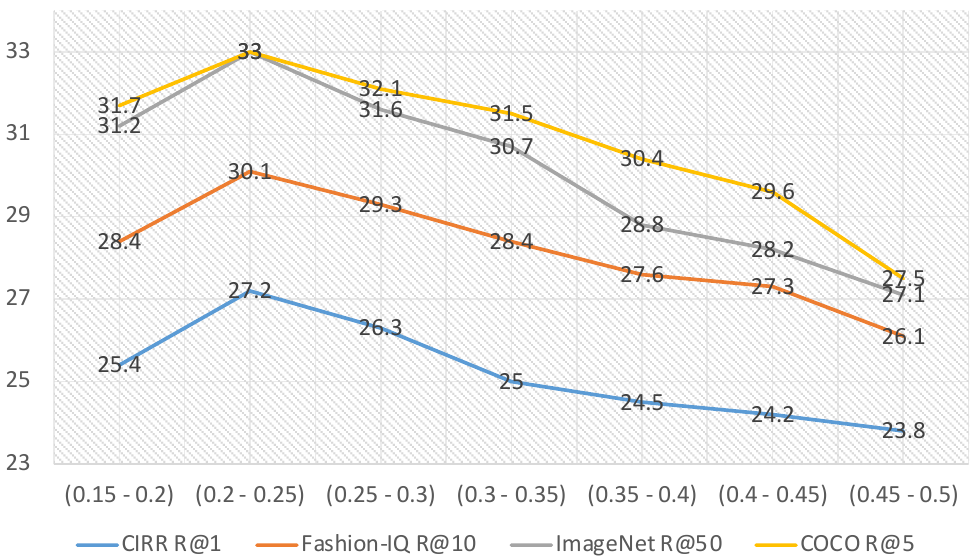}
    \caption{Analysis of the crop size for source/target view.}
    \vspace{-10pt}
    \label{fig:crop_ablation}
\end{figure}

\subsection{Analysis}
In this subsection, we provide detailed analyses of our design choices, efficiency, and common failure cases.

\noindent\textbf{Analysis of the Crop Size of World View.}
We analyze the influence of crop size for source and target views, as illustrated in Figure \ref{fig:crop_ablation}. A crop size range of $(0.15, 0.2)$ fails to learn sufficient target-relevant missing visual elements due to inadequate context in the source view. Increasing the crop size to $(0.45, 0.5)$ proves redundant, leading to excessive context overlap with the caption. We choose the crop size in the range $(0.2, 0.25)$, which gives the best result among different settings.

\begin{figure}[t]
    \centering
    \centering
    \includegraphics[width=1.0\linewidth]{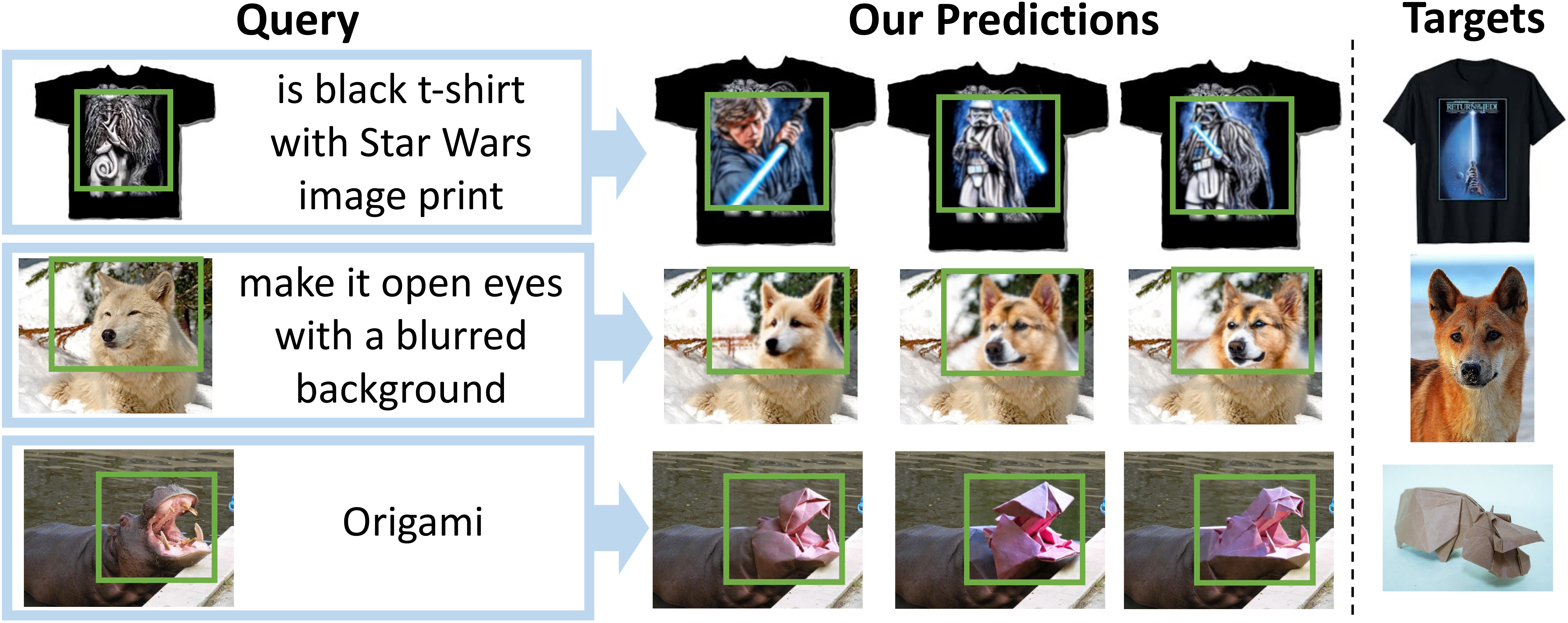}
    \caption{Visualization of our predictor representations. Green bounding boxes contain samples from a generative model decoding the output of our pretrained predictor.}
    \label{fig:pred_vis}
    \vspace{-8pt}
    
\end{figure}

\noindent\textbf{Visualization of Predictor Representations.} Following I-JEPA \cite{assran2023self}, we freeze our model and train a decoder following the RCDM framework \cite{bordes2021high}  to map the average pool of the predictor outputs back to pixel space. In Figure \ref{fig:pred_vis}, we show decoder outputs for various random seeds. The PrediCIR correctly predicts the target visual content missing in reference images guided by manipulation texts (\textit{e.g.,} Stars Wars print, open eyes, and origami style). For more details and samples, please refer to Appendix \ref{sec::more_predic_samples}. 

\noindent\textbf{Effectiveness and Efficiency Analysis.}
Our approach achieves significant improvements across six ZS-CIR tasks, with performance gains ranging from 1.73\% to 4.45\% Over SoTA models.  Due to our predictor design for prediction-based mapping, our model size is larger than the simple MLP mapping of Pic2Word. As a result, in the same setting, our training time (28 hours) is 6 hours longer than Pic2Word and 18 hours longer than SEARLE. Moreover, PrediCIR completes training 203 hours faster than the diffusion-based semi-supervised CompoDiff, achieving significant performance gains. Our inference time(0.03s) is only 0.01s slower than LinCIR and four times faster than CompoDiff (0.12s).

\begin{figure}[t]
    \centering
    \includegraphics[width=1.0\linewidth]{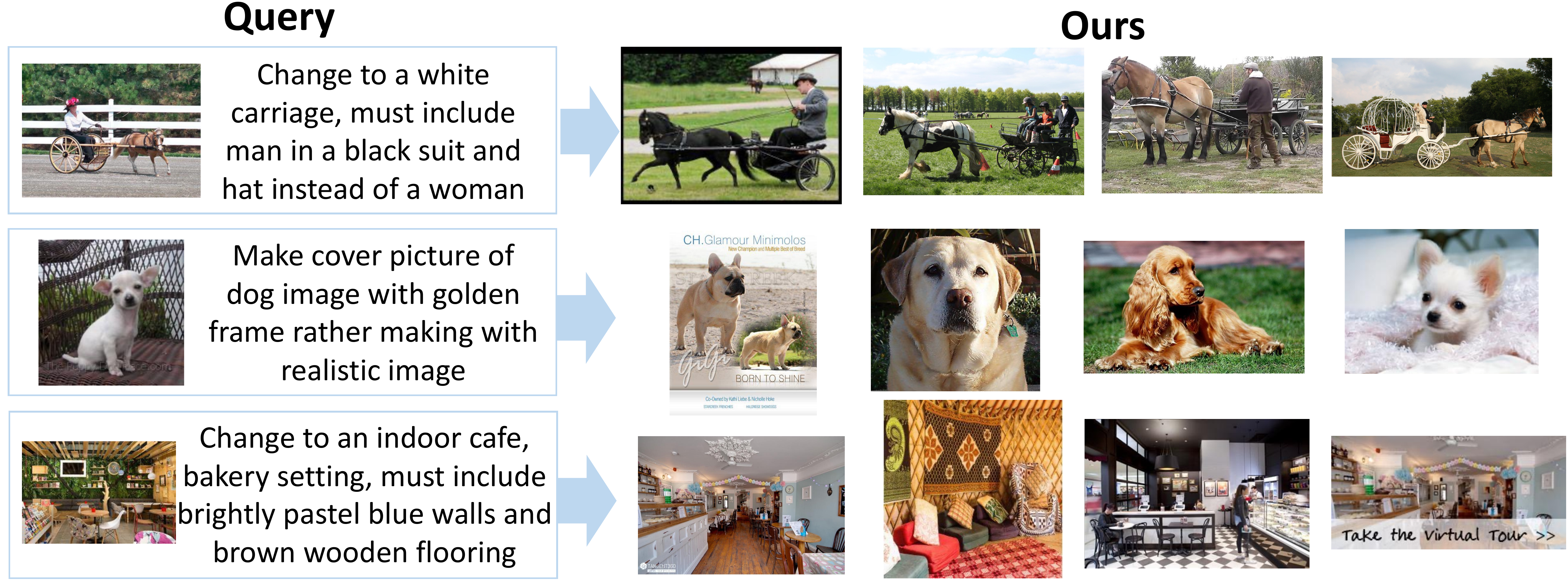}
    \caption{Visualization of common failure cases.}
    \label{fig:fail}
    \vspace{-10pt}
\end{figure}

\noindent\textbf{Discussion on Common Failure Case.}
Figure \ref{fig:fail} depicts PrediCIR's common failure cases, particularly with complex and redundant manipulation texts. Challenges include handling multiple objects and attributes (row 1), manipulating objects while converting image domains (row 2), and extensive concurrent manipulation of attributes and scenes (row 3). We believe these difficulties arise from the limitations of the CLIP language encoder in interpreting abstract or redundant intentions for retrieval.

\section{Conclusion}
\label{sec:conclusion}
In this paper, we propose a novel predicted-based image-to-word mapping method that leverages existing image-caption pairs to train a world model for predicting target visual content at latent space that is missing in reference guided by manipulation intention for accurate ZS-CIR. PrediCIR shows strong generalization ability and remarkably improves the best performance of existing approaches on six ZS-CIR tasks. It inspires the vision-to-language alignment mechanism and impacts diverse word modal applications.  How to design more lightweight and efficient models with high performance will be the future work.

\section*{Acknowledgements}
This work was supported by the Central Guidance for Local Special Project (Grant No. Z231100005923044).

{
    \small
    \bibliographystyle{ieeenat_fullname}
    \bibliography{main}
}
\newcommand{\methodNameNS}{\texttt{PredicCIR}}
\clearpage
\renewcommand{\thesection}{\Alph{section}}
\setcounter{section}{0}

\section{More Ablation Study}
\label{secc:more_ablation}
Table \ref{tab:more_ablation} presents additional ablation analyses for our PrediCIR model. \textbf{In models `1-3', we assessed the impact of varying crop sizes for constructing source and target views.} Using different crop sizes, unlike the consistent size in model `1', results in significant performance degradation. This decline is attributed to discrepancies in position embeddings between the source and target views, which complicate the model's ability to predict features spatially aligned with the reference image. \textbf{In models `4-6', we explored the effects of different aspect ratios.} Altering the aspect ratios, whether increasing (model `5') or decreasing (model `6'), led to an average performance decline of 2.30\% and 3.22\%, respectively, underscoring the sensitivity of model performance to aspect ratio adjustments. \textbf{In models `7-11', we further evaluated the impact of alternative solutions for key modules.} The results demonstrate that omitting our dynamic cropping strategy (model `8') or excluding reference image features in gating, which solely predicts the entire target image (model `9'), resulted in average performance reductions of 4.12\% and 3.50\%, respectively. This confirms the critical role of our strategies in maintaining model efficacy. Additionally, attempting to predict multiple target views from a single source view (model `10') also led to an average performance decline of 2.40\%, further validating the effectiveness of our targeted cropping strategy. Using a Faster R-CNN detector on CC3M for semantic-aware cropping (model `11'), which resulted in a 3.90\% performance decrease on CIRR and FashionIQ. While this strategy preserves object boundaries, it limits the diversity of training samples, particularly for fine-grained attribute manipulations like FashionIQ (drops by 6.15\%).  In contrast, our simple but effective random cropping strategy ensures richer and more variable training samples, benefiting the predictive world model despite possible inappropriate bboxs, which aligns with prior findings (\textit{e.g.,} MAE \cite{He_2022_CVPR}, I-JEPA \cite{lecun2022path,assran2023self}).

\begin{figure*}[htbp]
\vspace{30pt}
    \centering
    \centering
    \includegraphics[width=1.0\linewidth]{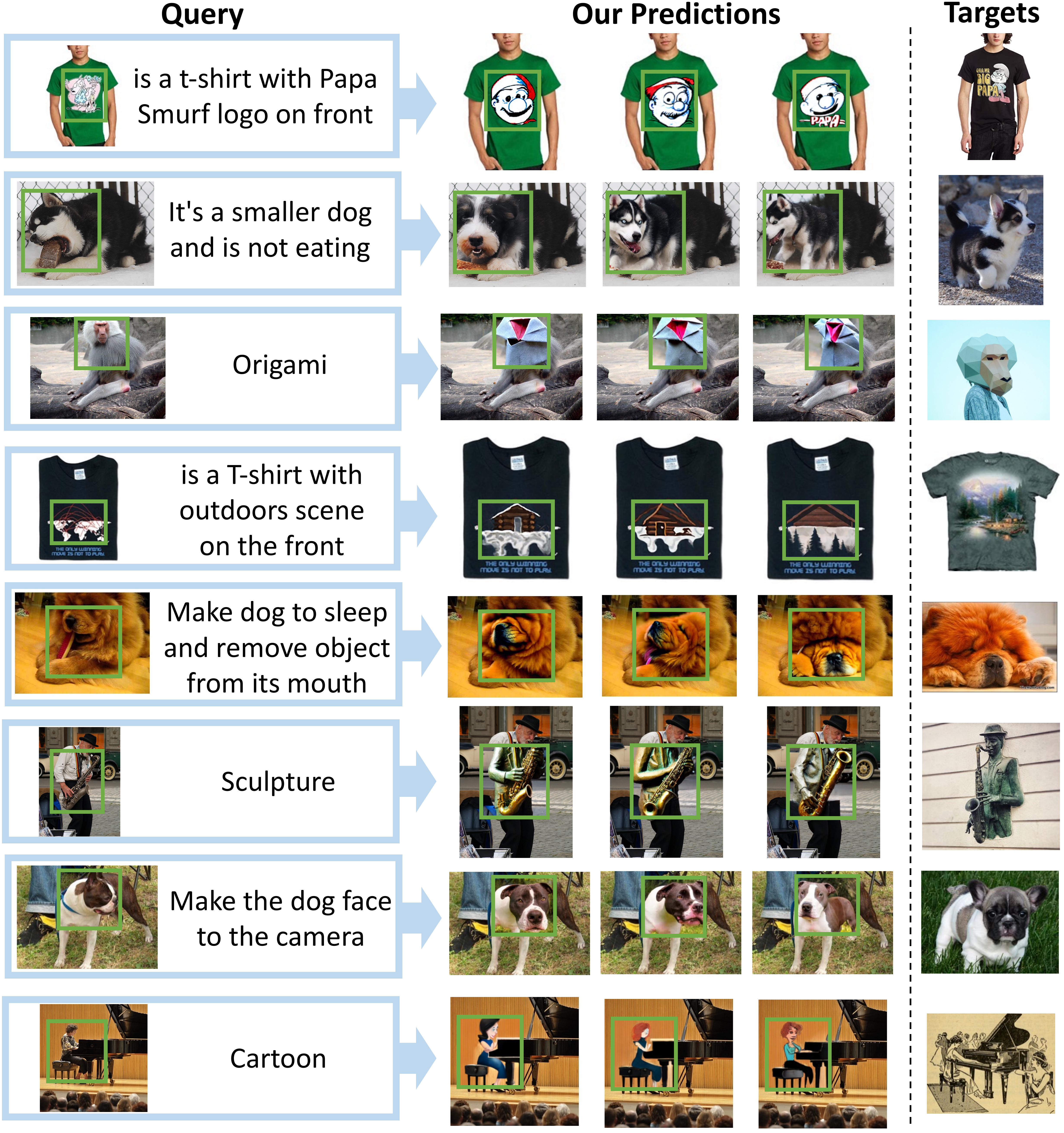}
    \caption{Visualization of our predictor representations. Green bounding boxes contain samples from a generative model decoding the output of our pretrained predictor.}
    \vspace{30pt}
    \label{fig:more_pred}
\end{figure*}

\begin{figure}[t]
    \centering
    \centering
    \includegraphics[width=1.0\linewidth]{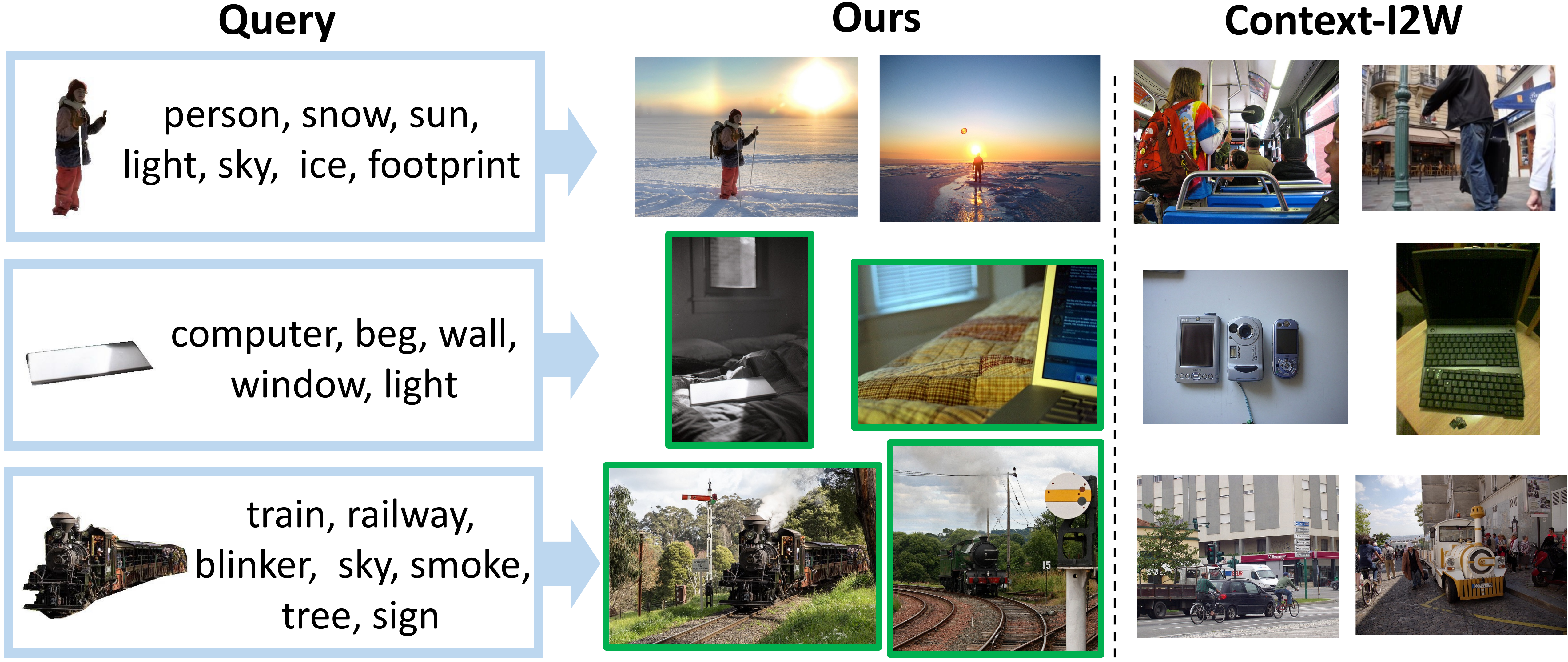}
    \caption{Retrieved results on the object composition task.}
    \label{fig:coco}
    \vspace{-5pt}
\end{figure}

\section{GeneCIS full results}
\label{sec::more_genecis}
In Table \ref{tab:more_genecis}, we report the full table of GeneCIS results. 

\section{Visualization of Predictor Representations}
\label{sec::more_predic_samples}

In Figure \ref{fig:more_pred}, we leverage the RCDM framework to visualize more samples of our PrediCIR's predicted target image feature into pixel space (Please refer to Section \ref{secc::RCDM_details}) for more details). The prediction effectively identifies the missing visual content in the reference images based on manipulation texts  (\textit{e.g.,} a Papa Smurf print, a dog not eating, a monkey in origami style, and a dog facing the camera). This pattern remains consistent, proving our predictor's ability to capture positional uncertainty and generate high-level visual elements (\textit{e.g.,}  objects, senses, attributes, and different details) with accurate poses. These results highlight the model's capacity for fine-grained visual content prediction, which is crucial for accurate ZS-CIR.

\begin{table}[t]
\centering
\scalebox{0.86}
{\footnotesize
\setlength{\tabcolsep}{1.5mm}
\begin{tabular}{llcclcc}
\toprule
\multicolumn{2}{l}{}   & \multicolumn{3}{c}{CIRR}                         & \multicolumn{2}{c}{Fashion-IQ} \\ \cmidrule(lr){3-5}\cmidrule(lr){6-7}
   & Methods           & R1   & R5   & R10                                & R10            & R50           \\ \cmidrule(lr){1-7}
\multicolumn{5}{l}{\textbf{Influence of different crop sizes for world view generation}} & & \\ 
1. & Source:$(0.2, 0.25)$, Target:$(0.2, 0.25)$ & 27.2 & 57.0 & \multicolumn{1}{l|}{70.2}          & 30.1           & 52.3          \\
2. & Source:$(0.15, 0.2)$, Target:$(0.2, 0.25)$ & 24.7 & 53.9 & \multicolumn{1}{l|}{66.2}          & 25.5           & 48.1          \\   
3. & Source:$(0.2, 0.25)$, Target:$(0.15, 0.2)$ & 25.3 & 54.8 & \multicolumn{1}{l|}{67.1}          & 26.8           & 49.4          \\ 
\multicolumn{5}{l}{\textbf{Influence of aspect ratios}} & & \\ 
4. & aspect ratios: $(0.75, 1.5)$  & 27.2 & 57.0 & \multicolumn{1}{l|}{70.2}          & 30.1           & 52.3          \\
5. & aspect ratios: $(1.0, 1.5)$ & 25.7 & 55.0 & \multicolumn{1}{l|}{67.5}          & 27.2           & 49.9          \\
6.& aspect ratios: $(0.75, 1.0)$             & 25.0 & 54.2 & \multicolumn{1}{l|}{66.4}          & 26.3           & 48.8          \\
\multicolumn{5}{l}{\textbf{Influence of different crop strategies}} & & \\ 
7. & single-blocks      & 27.2 & 57.0 & \multicolumn{1}{l|}{70.2}          & 30.1           & 52.3          \\
8. & w/o dynamic crop strategy     & 23.8  & 54.1 & \multicolumn{1}{l|}{66.8}          & 25.5           & 46.0          \\
9. & w/o source     & 24.5 & 54.7 & \multicolumn{1}{l|}{67.0}          & 25.9           & 47.2          \\   
10. & multi-blocks     & 25.6 & 55.2 & \multicolumn{1}{l|}{67.4}          & 27.2           & 49.4          \\
11. & semantic-aware crop strategy   & 24.7  & 55.1 & \multicolumn{1}{l|}{67.4}          & 24.8           & 45.3          \\   

\bottomrule
\end{tabular}}
\caption{More ablation study on CIRR and FashionIQ.}
\label{tab:more_ablation}
\vspace{-15pt}
\end{table}

\begin{table*}[t!]
    \centering  
    \resizebox{0.9\linewidth}{!}{
    \begin{tabular}{l|l|cccccccccccc|c}
    \toprule
    \multicolumn{2}{c|}{\textbf{GeneCIS $\rightarrow$}} & \multicolumn{3}{c}{Focus Attribute} & \multicolumn{3}{c}{Change Attribute} & \multicolumn{3}{c}{Focus Object} & \multicolumn{3}{c|}{Change Object} & \textbf{Average}\\
    \midrule
     Backbone & Method & R@1 & R@2 & R@3 & R@1 & R@2 & R@3 & R@1 & R@2 & R@3 & R@1 & R@2 & R@3 & R@1 \\
     \midrule
      \multirow{4}{*}{ViT-L/14}& SEARLE & 17.1 &	29.6 & 40.7 & 16.3 &	25.2 & 34.2 &	12.0 & \textbf{22.2} & 30.9 & 12.0 & 24.1 & 33.9 & 14.4\\
      & LinCIR & 16.9  & 30.0  & 41.5 & 16.2  & 28.0  & 36.8 & 8.3  & 17.4  & 26.2 & 7.4  & 15.7  & 25.0 & 12.2 \\
      & Context-I2W & 17.2  & 30.5  & 41.7 & 16.4  & 28.3  & \textbf{37.1} & 8.7  & 17.9  & 26.9 & 7.7  & 16.0  & 25.4 & 12.7 \\
    & PrediCIR(50\%) & 17.7 & 31.4 & 42.2 & 17.8 & 29.3 & 35.4 & 10.7 & 18.4 & 29.0 & 12.5 & 20.6 & 29.8 & 14.7 \\
      & \textbf{PrediCIR(100\%)} & \textbf{18.2}  & \textbf{31.9}  & \textbf{42.6} & \textbf{18.7}  & \textbf{30.4}  & 35.4 & \textbf{12.7}  & 19.0  & \textbf{31.2} & \textbf{16.9}  & \textbf{25.5}  & \textbf{34.1} & \textbf{16.6} \\
     \midrule
     & LinCIR & 19.1  & 33.0  & 42.3 & 17.6 & 30.2 & 38.1 & 10.1  & 19.1  & 28.1 & 7.9  & 16.3  & 25.7 & 13.7 \\
     ViT-G/14$^*$ & CompoDfff & 14.3   & 26.7   & 38.4  & 19.7  & 28.8  & 37.4  & 9.2   &  19.1  & 25.8  & 18.7   & 31.7   & 40.2 & 15.5 \\ 
     & \textbf{PrediCIR} & \textbf{19.3}  & \textbf{33.2}  & \textbf{42.7} & \textbf{19.9} & \textbf{30.7} & \textbf{38.9} & \textbf{12.8}  & \textbf{19.4}  & \textbf{32.3} & \textbf{18.9}  & \textbf{32.2}  & \textbf{40.6} & \textbf{18.7} \\
    \bottomrule
    \end{tabular}}    
    \caption{\textbf{Comparison on GeneCIS Test Data.} PrediCIR is able to significantly outperform adaptive methods across all Fashion-IQ sub-benchmarks, with its inherent modularity allowing for further simply scaling to achieve additional large gains. ($^*$) OpenCLIP weights \cite{openclip}.}
    \label{tab:more_genecis}
\end{table*}

\section{More Qualitative Experiment on COCO}
In the object composition experiments, PrediCIR significantly outperforms the current SoTA model by an average of 3.60\%. These results underscore the remarkable effectiveness of our TCP module in predict missing objects relevant to manipulation text, which facilitates the combination of multiple objects, as shown in Figure \ref{fig:coco}.

\section{Algorithm of Prediction-based Word Mapping Process.} 

Algorithm \ref{alg:algorithm_sampler} presents the pseudo-code for our prediction-based image-to-word mapping process. We initiate the process by creating mask tokens for a target block. The mask tokens are
parameterized by a shared learnable vector with an added
positional embedding. These mask tokens are designed to predict the visual content missing in the reference image. These mask tokens are subsequently fed into a narrow Transformer architecture, which incorporates source local features and the action with manipulation intent to perform self-attention. To achieve a dynamic ratio during the fusion of source and predict embeddings, we utilize a tanh-gating mechanism \cite{hochreiter1997long}.

\section{Review of Image World Model}

\subsection{JEPA Framework Overview}
The Image World Model (IWM) \cite{garrido2024learningleveragingworldmodels}. builds upon the Joint Embedding Predictive Architecture (JEPA) framework~\citep{lecun2022path}, as utilized in approaches like I-JEPA~\citep{assran2023self}. In JEPA-based methods, representations are learned by predicting the effect of transformations applied to an image in a latent space. This is achieved by conditioning the predictor on transformation parameters, allowing it to infer the relationship between source and target representations effectively. 

Unlike contrastive methods that aim for invariance to data augmentations, JEPA frameworks preserve semantic information through latent inpainting, enabling the predictor to model transformations explicitly. By working in the latent space, JEPA removes redundant or hard-to-predict details, improving representation quality without focusing on pixel-level reconstruction~\citep{chen2024deconstructing}. These features make JEPA a powerful tool for learning representations that are both semantically meaningful and capable of generalization.

\subsection{Image World Model (IWM)}
IWM extends the JEPA framework to learn robust and reusable world models. The predictor in IWM serves as the instantiation of the world model, capable of applying transformations in latent space. Unlike invariant predictors, which disregard transformation details, IWM learns equivariant representations by conditioning on transformation parameters \cite{garrido2024learningleveragingworldmodels}.

The training process begins with the generation of source ($x$) and target ($y$) views from a given image $I$. Target views are created by applying random augmentations such as horizontal flips, cropping, and color jitter, ensuring the target retains as much semantic information as possible. In contrast, source views are derived from the target by introducing additional transformations, including grayscale, blur, solarization, and masking inspired by I-JEPA. These transformations enforce the predictor to learn transformation-aware latent representations.

\begin{algorithm}[tb]
\caption{Prediction-based Word Mapping process.}
\label{alg:algorithm_sampler}  
\textbf{Input}: batch of source image features $\boldsymbol{V}_x=  \{\boldsymbol{v}_{x_i}\}_{i=1}^m$, where $\boldsymbol{v}_{x_1}$ is the global source feature $\boldsymbol{v}_{x_g}$, batch of action \( a_{\boldsymbol{x}{\rightarrow \boldsymbol{y}}} \) with manipulation intent, $N_{layer}$.
\\
\textbf{Parameter}:  mask tokens \( m_a\), 
parameterized by a shared learnable vector $\boldsymbol{x}\in{\mathbb{R}^{d\times{1}}}$ with an added
positional embedding, 8-heads attention layer $Attn$, 3-layers FC layers $f_M$, $gate_\alpha$. \\
\textbf{Output}: pseudo token $S_*$
\begin{algorithmic}[1] 
\STATE Initialize $m_a\in{\mathbb{R}^{d\times{n}}}$, $Attn$, $f_M$ randomly. 
\STATE Let $\boldsymbol{X}^i_{att}= [a_{\boldsymbol{x}{\rightarrow \boldsymbol{y}}}, \{\boldsymbol{v}_{x_i}\}_{i=2}^m$, \(m_a\)],$t=1$
\WHILE{$t \le N_{layer}$}
\STATE $\boldsymbol{X}^{i+1}_{att} = \boldsymbol{X}^{i}_{att} + Attn_t$(q=$\boldsymbol{X}^{i+1}_{att}$, k=$\boldsymbol{X}^{i+1}_{att}$, v=$\boldsymbol{X}^{i+1}_{att}$)
\STATE $\boldsymbol{X}^{i+1}_{att} = \boldsymbol{X}^{i+1}_{att} + f_{M_t}(\boldsymbol{X}^{i+1}_{att})$
\STATE $t = t + 1$
\ENDWHILE

$S_*$ = $f_{M_{s}}$($\boldsymbol{v}_{x_g}) + tanh(gate_\alpha) \cdot avg$($f_{M_{p}}(\boldsymbol{X}_{out}))$

\STATE \textbf{return} $S_*$
\end{algorithmic}
\end{algorithm}

\paragraph{Transformation Encoding.} The transformation parameters $a_{x \rightarrow y}$ encode the differences between source and target views, including augmentation details such as color jitter and destructive transformations. These parameters serve as input to the predictor, allowing it to model the transformations explicitly.

\paragraph{Latent Prediction.} The source and target views are processed by an encoder $f_\theta$ and its exponential moving average (EMA) $f_\theta^\text{EMA}$ to obtain latent representations $z_x$ and $z_y$. The predictor $p_\phi$ is conditioned on the source embedding, transformation parameters, and masked token positions to predict the target representation $\hat{z_y}$. The learning objective minimizes the $L2$ distance between the predicted $\hat{z_y}$ and the actual target $z_y$ over masked regions:
\begin{equation*}
    L(x,y) = \sum_{i\in M_x^C}\| p_\phi\left(f_\theta(x),a_{x\rightarrow y},m_a \right)_i
    - f_\theta^\text{EMA}(y)_i \|_2^2.
\end{equation*}
\paragraph{Architecture.} The encoder of IWM adopts the ViT architecture~\cite{DBLP:journals/corr/abs-2010-11929}, while the predictor uses a similar structure with modified depth and embedding dimensions. IWM instances are denoted as $\text{IWM}_{X,Y}^Z$, where $X$ is the predictor depth, $Y$ its embedding dimension, and $Z$ specifies its capability, such as "Equi" for equivariant models.

\subsection{The Reusability of IWM}
IWM not only enhances representation learning but also enables effective downstream task adaptation. Finetuning the learned world model alongside the frozen encoder significantly improves task performance with minimal additional cost. Furthermore, inspired by instruction tuning~\cite{wei2021finetuned}, IWM can be adapted for multi-task learning, demonstrating its efficiency and versatility compared to traditional methods. This highlights the importance of incorporating the world model into inference processes, rather than discarding it after pretraining.

\section{More Implementation Details}

For training PrediCIR,  We adopt  ViT-B/32 and ViT-L/14 CLIP \cite{radford2021learning} pre-trained on 400M image-text paired data. The crop sizes and aspect ratios of random cropped images and blocked target images are the same, in the range of $(0.2, 0.25)$ and $(0.75, 1.5)$, respectively (ablation in the supplementary). For training PrediCIR, we utilize the Conceptual Caption dataset \cite{DBLP:conf/acl/SoricutDSG18}, which comprises 3M images.  Our predictor is designed as a lightweight (narrow) ViT architecture. Specifically, the number of self-attention blocks is $12$ with $384$ dimensional embeddings. To improve training stability, we initialize the learnable scalar of tanh-gating to 0 \cite{bachlechner2021rezero}. We employ AdamW \cite{loshchilov2018decoupled} with a learning rate of $1\times10^{-5}$, weight decay of $0.1$, and a linear warmup of $10000$ steps. The batch size is $1024$.  All models are trained on \(4\) NVIDIA A100 (80G) GPUs. For training Pic2Word, SEARLE, Context-I2W, and LinCIR, we utilized their official code for training, and hyper-parameters were kept consistent with those reported in their respective papers.  To ensure reliable results, we report the performance averaged over three trials.

\subsection{RCDM Visualizations Details.} 
\label{secc::RCDM_details}
In Figure 7 of our main paper and Figure \ref{fig:more_pred}, to visualize the representations of a pre-trained neural network in pixel space, we follow I-JEPA \cite{assran2023self}, freeze our PrediCIR, and train a decoder following the RCDM framework \cite{bordes2021high}. The RCDM framework trains a decoder network $h_\omega$, comprising a generative diffusion model, to reconstruct an image $\vx$ from the representation vector of that image $\vs_x$ and a noisy version of that image $\hat{\vx} \coloneqq \vx + \epsilon$, where $\epsilon$ is an additive noise vector. Concretely, the decoder objective is to minimize the loss function $\lVert h_\omega(\hat{\vx}, \vs_x) - \epsilon \rVert$. We train each RCDM network for 350,000 iterations using the default hyperparameters. After training the decoder, one can subsequently feed the representation vector of an unseen test image $\vs_y$ into the decoder along with various random noise vectors to generate several pixel-level visualizations of the representation, thus providing insight into the features captured in the representations of the pre-trained network. Qualities that are common across samples represent information that is contained in the representation. On the other hand, qualities that vary across samples represent information that is not contained in the representations. 

\subsection{More Inference Details}
\label{secc::more_inference}
\noindent \textbf{(1) Domain conversion}. This setup evaluates the ability to compose real images and domain information to retrieve corresponding domain-specific images. We utilize ImageNet \cite{deng2009imagenet} and ImageNet-R \cite{Hendrycks_2021_ICCV}, which comprises 200 classes with diverse domains and has domain annotations. Following Pic2Word, we pick cartoon, origami, toy, and sculpture as the evaluation target to avoid noise in the annotations. With this selection, we have 16,983 images as candidates. In the evaluation, given the real image from ImageNet and target domain names, we compose the query following the procedure in (a) in the Inference section. \textit{e.g.,} \texttt{a cartoon of [*]}.

\noindent \textbf{(2) Object/Attribute composition}.  We evaluate the GeneCIS \cite{vaze2023genecis} test split and the validation split (5000 images) of COCO \cite{10.1007/978-3-319-10602-1_48},, which dataset contains images with corresponding lists of object classes and instance mask of query images. Following Pic2Word, we randomly crop one object and mask its background using its instance mask to create a query for each image. The list of object classes is used as text specification. Given the reference image and class list, we compose a query by following (b) in the Inference section. \textit{e.g.,} \texttt{a photo of [*], [cat] and [dog]}.

\noindent \textbf{(3) Object/scene manipulation by text description}. In this setup, a reference image is provided alongside a text description containing instructions for manipulating either an object or the background scene depicted in the reference image. This composition of the reference image and text description enables the retrieval of manipulated images. We evaluate the test split of CIRR \cite{Liu_2021_ICCV} and CIRCO \cite{baldrati2023zero} using the standard evaluation protocol following previous works \cite{Saito_2023_CVPR,baldrati2023zero,tang2023contexti2w}, and query texts are composed following the procedure \texttt{a photo of [*], [sentence]}.

\noindent \textbf{(4) Attribute manipulation}. We employ Fashion-IQ \cite{Wu_2021_CVPR}, which includes various modification texts related to image attributes. These attribute manipulations are given as a sentence. As with CIRR, we adopt the standard evaluation protocol and create query texts following the procedure \texttt{a photo of [*], [sentence]}. In evaluation, we employ the validation set, following previous works \cite{Baldrati_2022_CVPR,Saito_2023_CVPR,baldrati2023zero,tang2023contexti2w}.
\end{document}